\documentclass[10pt,twocolumn,letterpaper]{article}

\usepackage{cvpr}              
\usepackage[dvipsnames]{xcolor}

\definecolor{cvprblue}{rgb}{0.21,0.49,0.74}
\usepackage[pagebackref,breaklinks,colorlinks,citecolor=cvprblue]{hyperref}
\usepackage{multirow}
\usepackage{pifont}
\definecolor{darkgreen}{rgb}{0.0, 0.5, 0.0}

\newcommand{\cmark}{\textcolor{darkgreen}{\ding{51}}}  
\newcommand{\xmark}{\textcolor{red}{\ding{55}}}        

\usepackage[dvipsnames]{xcolor}
\usepackage{listings}

\title{EvalMuse-40K: A Reliable and Fine-Grained Benchmark with Comprehensive Human Annotations for Text-to-Image Generation Model Evaluation}

\author{Shuhao Han$^{1,2 *}$\quad
Haotian Fan$^{2}$\quad
Jiachen Fu$^{1}$\quad
Liang Li$^{2}$\quad
Tao Li$^{2}$\quad\\
Junhui Cui$^{2}$\quad
Yunqiu Wang$^{2}$\quad
Yang Tai$^{2}$\quad
Jingwei Sun$^{2}$\quad
Chunle Guo$^{1,3}$\quad
Chongyi Li$^{1,3}$\quad
\vspace{0.1em}\\
$^1$VCIP, CS, Nankai University\quad 
$^2$ByteDance Inc\quad 
$^3$NKIARI, Shenzhen Futian\quad \\
\\
\url{https://shh-han.github.io/EvalMuse-project/}
}

\begin{document}
\maketitle
\renewcommand{\thefootnote}{\fnsymbol{footnote}}
\footnotetext[1]{Intern in Media Evaluation Lab, ByteDance Inc}
\begin{abstract}
Text-to-Image (T2I) generation models have achieved significant advancements. 
Correspondingly, many automated metrics emerge to evaluate the image-text alignment capabilities of generative models.
However, the performance comparison among these automated metrics is constrained by the limited scale of existing datasets.
Additionally, these datasets lack the capacity to assess the performance of automated metrics at a fine-grained level.
In this study, we contribute an EvalMuse-40K benchmark, gathering 40K image-text pairs with fine-grained human annotations for image-text alignment-related tasks.
In the construction process, we employ various strategies such as balanced prompt sampling and data re-annotation to ensure the diversity and reliability of our benchmark.
This allows us to comprehensively evaluate the effectiveness of image-text alignment metrics for T2I models.
Meanwhile, we introduce two new methods to evaluate the image-text alignment capabilities of T2I models: FGA-BLIP2 which involves end-to-end fine-tuning of a vision-language model to produce fine-grained image-text alignment scores and PN-VQA which adopts a novel positive-negative VQA manner in VQA models for zero-shot fine-grained evaluation.
Both methods achieve impressive performance in image-text alignment evaluations. 
We also apply our methods to rank existing AIGC models, providing results that can serve as a reference for future research and foster the development of T2I generation.
\end{abstract}    
\section{Introduction}
\begin{table*}[!ht]
    \centering
        \caption{In comparison to existing T2I model evaluation benchmarks, EvalMuse-40K collects a large number of human annotations (Ann.). Furthermore, EvalMuse-40K offers fine-grained annotations at the element (Ele.) level and categorizes these elements into different skills in image-text alignment. Additionally, EvalMuse-40K includes annotations for structural problems in generated images. To ensure reliable evaluation of automated metrics, we also randomly generate image-text pairs from 20 different T2I models.}
        \vspace{-2mm}
    \label{tab:benchmark}
    \begin{tabular}{l|ccc|ccc|c|c}
    \hline
        \multirow{2}{*}{Benchmark} & \multicolumn{3}{c|}{Dataset Size} & \multicolumn{3}{|c|}{Alignment} & Faithfulness & T2I Models \\ \cline{2-9}
        ~ & Prompt  & Image & Ann. & Likert & Ele.-Ann. & Ele.-Category& Structure  & Num.\\ \hline
        PartiPrompt~\cite{yu2022scaling} & 1.6K & - & - & \xmark & \xmark & \xmark & \xmark & - \\ 
        DrawBench~\cite{saharia2022photorealistic} & 200 & - & - & \xmark & \xmark & \xmark & \xmark & -  \\ 
        T2I-CompBench~\cite{huang2023t2i} & 6K & - & - & \xmark & \xmark & \cmark & \xmark & -  \\ 
        TIFA160~\cite{hu2023tifa} & 160 & 800 & 1.6K & \cmark & \xmark & \cmark & \xmark & 5 \\ 
        GenAI-Bench~\cite{li2024genai} & 1.6K & 9.6K & 28.8K & \cmark & \xmark & \xmark & \xmark & 6 \\ 
        Gecko~\cite{wiles2024revisiting} & 2K & 8K & 108K & \cmark & \cmark & \xmark & \xmark &  4 \\
 EVALALIGN~\cite{tan2024evalalign}& 3K& 21K& 132K& \cmark& \xmark& \cmark& \cmark & 8 \\ 
        \textbf{EvalMuse-40K (ours)} & \textbf{4K} & \textbf{40K} & \textbf{1M} & \cmark & \cmark & \cmark & \cmark & \textbf{20}  \\ \hline
    \end{tabular}

\end{table*}

Recently, advanced Text-to-Image (T2I) models~\cite{li2024hunyuan,peebles2023scalable,esser2024scaling,podell2023sdxl,rombach2022high,saharia2022photorealistic,yu2022scaling} are capable of generating numerous impressive images. 
However, these models may still generate images that fail to accurately match the input text, such as inconsistency in quantities~\cite{kirstain2023pick,xu2024imagereward,wu2023human}. 
Given the high cost and inefficiency of manual evaluation, developing a reliable automatic evaluation metric and corresponding benchmark is vital.
They can effectively evaluate the performance of existing models and provide guidance for improvements in future models.

Since traditional evaluation metrics, such as FID~\cite{heusel2017gans} and CLIPScores~\cite{hessel2021clipscore}, are not well-suited for assessing the consistency of T2I models, recent works~\cite{hu2023tifa,wiles2024revisiting,yarom2024you} have explored the construction of new evaluation metrics. 
These methods introduce diverse evaluation approaches leveraging various multi-modal models.
For example, VQAScore~\cite{li2024evaluating} is constructed by asking the Visual Question Answering (VQA) model ``Does this figure show \{text\}?" and obtaining the likelihood of answer ``Yes" as the alignment score.
However, such a simple VQA manner cannot handle the fine-grained matching problem well. 
TIFA~\cite{hu2023tifa} and VQ2~\cite{yarom2024you} decompose the prompt into multiple elements, formulate related questions for each, and then average the answers to generate a final alignment score.
These methods allow for fine-grained evaluation of image-text alignment but are also limited by the performance of the VQA model. 
Moreover, the relationship between element-based alignment scores and overall human preferences has yet to be explored.
Therefore, to better explore the performance of existing T2I evaluation methods, we contribute a new benchmark, \textbf{EvalMuse-40K}, featuring fine-grained human annotations of image-text pairs.

EvalMuse-40K includes 4K prompts, 40K image-text pairs, and more than 1M fine-grained human annotations.
To ensure the diversity of prompts, EvalMuse-40K includes 2K real prompts and 2K synthetic prompts, where the real prompts are sampled from DiffusionDB~\cite{wang2022diffusiondb}.
We categorize the real prompts from multiple dimensions and use the MILP~\cite{vonikakis2016shaping} strategy to ensure the category balance of the final sampled prompts.
The synthesized prompts are then constructed for specific skills in image-text alignment, such as quantity and location. 
By synthesizing specific prompts and sampling from real prompts, EvalMuse-40K is able to accurately evaluate specific model problems while also providing a more comprehensive assessment of model performance in real-world scenarios.
For fine-grained evaluation, we use a large language model for the elemental splitting of prompts and question generation, and to increase the diversity of the generated images, we generate images using a variety of T2I models.
Compared with the previous benchmark (see Tab.~\ref{tab:benchmark} for details), EvalMuse-40K not only scores image-text alignment but also performs more fine-grained annotations for elements split from the prompts.
Additionally, we annotate different types of structural problems that may appear in generated images.
This comprehensive and fine-grained human evaluation in EvalMuse-40K enables an analysis of current automated T2I model evaluation metrics and their correlation with human preferences, ultimately supporting the improvement and development of the T2I evaluation system.

In addition to the comprehensive benchmark, we introduce two methods, FGA-BLIP2 and PN-VQA, for image-text alignment evaluation.
FGA-BLIP2 uses vision-language models to jointly fine-tune the image-text alignment scores and element-level annotations so that the models can output the overall scores and determine whether the generated images match the elements in the prompt. 
Additionally, we employ a variance-weighted optimization strategy to account for prompts’ ability to generate diverse images.
PN-VQA constructs specific positive-negative question-answer pairs, allowing MLLMs to evaluate the fine-grained alignment of image-text pairs. 
With these two methods, we can obtain a scoring model aligned with human preferences and perform fine-grained alignment analysis and evaluation.

Finally, we select some representative prompts to regenerate images and analyze existing AIGC models using our proposed evaluation methods. 
Our data, models, code, and results serve as valuable resources to support further research.

\textbf{To summarize, our contributions are listed as follows.}
\begin{itemize}
    \item EvalMuse-40K collects a large number of human annotations and ensures a more robust evaluation of automated metrics by using a well-balanced set of prompts and a diverse range of generative models.
    \item EvalMuse-40K categorizes elements during fine-grained annotation, enabling the evaluation of existing automated metrics' accuracy in assessing specific skills at a fine-grained level.
    \item We enable fine-grained evaluation by fine-tuning a scoring model, FGA-BLIP2, achieving impressive performance across multiple datasets.
    \item We introduce a new automated metric, PN-VQA, which employs positive-negative questioning to enhance the correlation with human ratings.
\end{itemize}

\section{Related Work}
\begin{figure*}
    \centering
    \includegraphics[width=1\textwidth]{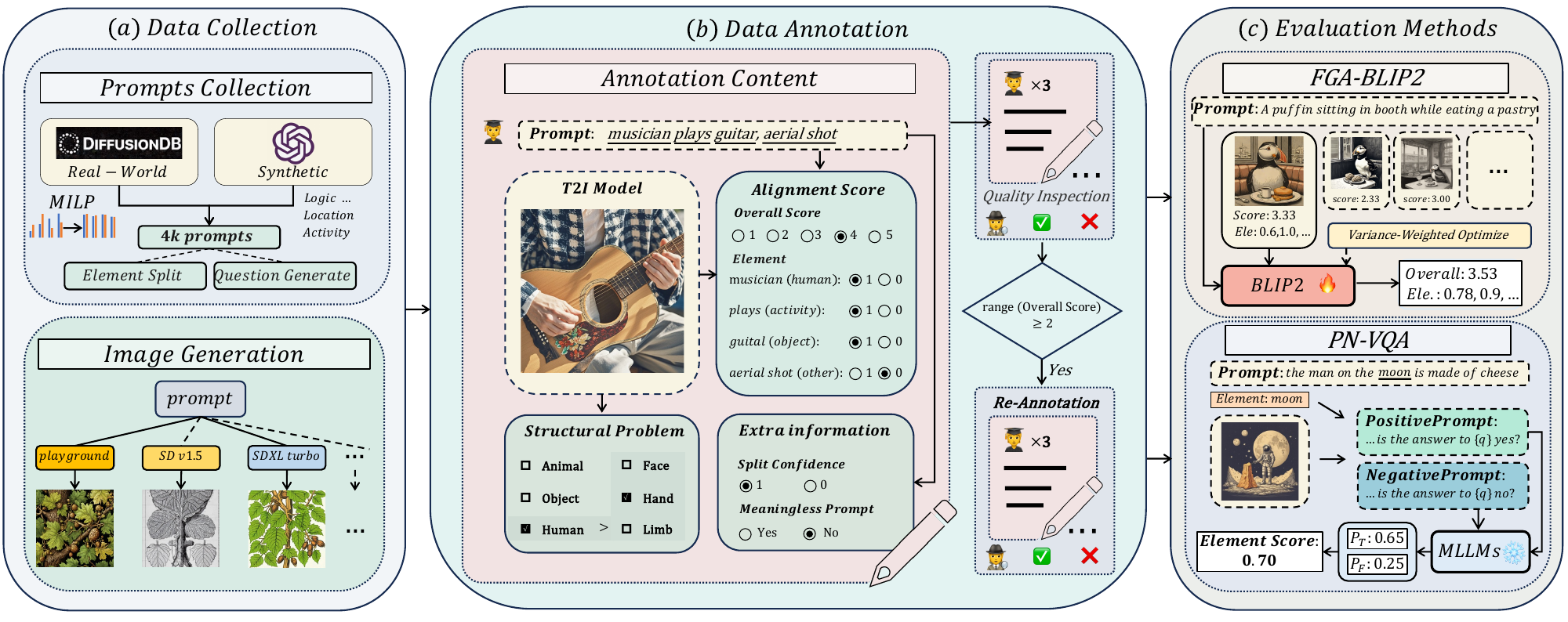}
    \caption{Overview construction of EvalMuse-40K, consisting of (a) data collection, (b) data annotation, and (c) evaluation methods. (a) We collect 2K real-world prompts and 2K synthetic prompts, using the MILP~\cite{vonikakis2016shaping} sampling strategy to ensure category balance and diversity. Prompts are further divided into elements, and corresponding questions are generated. We also use various T2I models to generate images. (b) During data annotation, we label the fine-grained alignment levels of image-text pairs, structural problems of the generated images, and some extra information. For annotated data with large differences in overall alignment scores, we perform re-annotation to ensure the reliability of the benchmark. (c) We propose two effective image-text alignment evaluation methods: one is FGA-BLIP2, using a fine-tuned vision-language model for direct fine-grained scoring of image-text pairs, and another is PN-VQA, adopting positive-negative VQA manner for zero-shot evaluation.} 
     \vspace{-2mm}
    \label{fig:benchmark}
\end{figure*}

\noindent
\textbf{Image-Text Alignment Benchmarks.}
Many different benchmarks have been proposed to evaluate the image-text alignment of T2I models. 
Early benchmarks are small-scale and mostly rely on captions from existing datasets like COCO ~\cite{cho2023dall,hu2023tifa,lin2014microsoft,ramesh2022hierarchical}, focusing on limited sample skills.
Other benchmarks (e.g., HPDv2~\cite{wu2023human} and Pick-a-pic~\cite{kirstain2023pick}) use side-by-side model comparisons to evaluate the quality of the generated images. 
Recently, benchmarks like DrawBench~\cite{saharia2022photorealistic}, PartiPrompt~\cite{yu2022scaling}, and T2I-CompBench~\cite{huang2023t2i} have introduced a set of prompts and focused on evaluating specific skills of generative models, including counting, spatial relationships, and attribute binding. 
Moreover, some benchmarks (e.g., GenAI-Bench~\cite{li2024genai} and  RichHF-18K~\cite{liang2024rich}) provide human annotations on image-text align scores to validate the relevance of automated metrics with human preference. 
For fine-grained evaluation, benchmarks like TIFA~\cite{hu2023tifa} and SeeTRUE~\cite{yarom2024you} extract elements from prompts and generate corresponding questions. 
%
However, prompts in current benchmarks often focus on specific T2I model skills, which differ from prompts used by real users.  
The proposed EvalMuse-40K addresses this by containing 2K real prompts sampled from DiffusionDB~\cite{wang2022diffusiondb} and 2K synthetic prompts for specific skills, enabling a more comprehensive evaluation of image-text alignment in image generation.
Additionally, it provides large-scale human annotations on both overall and fine-grained aspects of image-text alignment, allowing deeper analysis of how well current evaluation metrics align with human preference.
 
\noindent
\textbf{Automated Metrics for Image-Text Alignment.}
Image generation models initially use FID~\cite{heusel2017gans} to calculate the distributional differences between generated images and validation images. 
Besides, IS~\cite{salimans2016improved} and LPIPS~\cite{zhang2018unreasonable} have been used to evaluate the quality of generated images. 
However, since these metrics do not effectively evaluate image-text alignment, recent work primarily reports CLIPScore~\cite{hessel2021clipscore}, which measures the cosine similarity of text and image's features.
With the strong performance of BLIP2~\cite{li2023blip}, BLIP2Score has also been adopted as an alignment metric similar to CLIPScore.
Human preference models (e.g., ImageReward~\cite{xu2024imagereward}, PickScore~\cite{kirstain2023pick}, and VNLI~\cite{yarom2024you}) improve assessment capabilities by using a vision-language model like CLIP, fine-tuned on large-scale human ratings.
However, these models primarily rely on side-by-side image comparison annotations to learn human preferences, making it difficult to obtain accurate alignment scores and perform fine-grained evaluations.
We address this by using a vision-language model to directly predict both overall and fine-grained alignment scores for image-text pairs. 
For fine-grained evaluation, TIFA~\cite{hu2023tifa}, VQ2~\cite{yarom2024you}, and Gecko~\cite{wiles2024revisiting} split prompts into elements and generate questions to assess fine-grained issues in generated images using VQA models. 
However, these generated questions can sometimes diverge from the original prompt’s content. 
To address this, we design a new positive-negative question template that provides VQA models with more accurate prompt context, improving alignment with human preference.

\section{EvalMuse-40K Benchmark}

In this section, we detail EvalMuse-40K, a reliable and fine-grained benchmark with comprehensive human annotations for T2I evaluation. We present the overview construction of EvalMuse-40K in Fig.~\ref{fig:benchmark}.
In the construction process, we employ various strategies to ensure the diversity and reliability of our benchmark.
Next, we introduce the construction process of EvalMuse-40K from three aspects: prompt collection, image generation, and data annotation followed by a statistical analysis of the data.

\subsection{Prompts Collection}
To better evaluate the T2I task, EvalMuse-40K collects 2K real prompts from DiffusionDB~\cite{wang2022diffusiondb} and 2K synthetic prompts for specific skills. 
Below, we outline the process of collecting real and synthetic prompts and performing element splitting and question generation.

\noindent \textbf{Real Prompts from Community.} 
Most prompts used by current benchmarks are generated from templates with LLMs or manually crafted, with only a small portion originating from real-world users.
This results in a gap between model evaluation and actual user needs.
To address this, we sample 2K real prompts from DiffusionDB~\cite{wang2022diffusiondb}, which contains 1.8M prompts specified by real users.

To be specific, we first randomly sample 100K prompts from DiffusionDB as the source dataset. To ensure diversity in the final sampled prompts, we classify the prompts in four aspects: subject category, logical relationship, image style, and BERT~\cite{devlin2018bert} embedding score. 
The specific categories of each aspect are shown in the supplementary material. 
Except for the BERT embedding score, which is computed, the other aspects are categorized using GPT-4~\cite{achiam2023gpt}, and a prompt can belong to more than one category.
After labeling the 100K prompts, we improve the data shaping method proposed in~\cite{vonikakis2016shaping} by using Mixed Integer Linear Programming (MILP) to ensure that the sampled data is approximately uniformly distributed across each category.
The specific sampling strategy and the distribution of the sampled data are shown in the supplementary material.

\noindent \textbf{Synthetic Prompts for Various Skills.}
To comprehensively evaluate the generative models' skills, we employ specific templates and a rich corpus to generate 2K synthetic prompts using GPT-4.
These prompts are divided into six categories: (1) Object Count; (2) Object Color and Material; (3) Environment and Time Setting; (4) Object Activity and Perspective Attributes; (5) Text Rendering; and (6) Spatial Composition Attributes. 
For each category, we use different templates and GPT-4 to generate reasonable and natural prompts. 
The specific generation methods and details are provided in the supplementary material.

\noindent \textbf{Element Splitting and Question Generation.} 
To achieve fine-grained annotation and evaluation, we perform element splitting and question generation on the 4K collected prompts. 
In contrast to the word-level annotation used in RichHF~\cite{liang2024rich} and Gecko~\cite{wiles2024revisiting}, we split the prompt into fine-grained elements, which increases the consistency of the annotations across different humans.
In addition, categorizing each element allows us to examine the model's capabilities in various aspects at a fine-grained level.
We adopt the element categorization strategy from TIFA~\cite{hu2023tifa} but refine the question generation process by dividing it into two steps: first, splitting the prompts into individual elements, and then generating specific questions for each element.
We generate only simple judgment questions with yes or no answers, making the question generation process more controllable and the evaluation more explainable. 
The generated questions are also filtered and regenerated to ensure that each element has a corresponding question. 
The templates used for element splitting and question generation are shown in the supplementary material.

\subsection{Image Generation}
To ensure the diversity of the generated images, we select 20 different types and versions of diffusion-based generative models, considering the strong performance of diffusion models~\cite{ho2020denoising}.
Meanwhile, to ensure the differentiation among the selected models, we classify the selected models into four groups: (1) basic stable diffusion~\cite{rombach2022high} models, such as SD v1.2, SD v1.5, SD v2.1; (2) advanced open-source generative models, such as SDXL~\cite{podell2023sdxl}, SSD1B~\cite{gupta2024progressive}, SD3~\cite{esser2024scaling}, HunyuanDiT~\cite{li2024hunyuan}, Kolors~\cite{kolors}, PixArt-$\alpha$~\cite{chen2023pixart}, PixArt-$\Sigma$~\cite{chen2024pixart}, IF~\cite{deep2023if}, Kandinsky3~\cite{arkhipkin2023kandinsky}, Playground v2.5~\cite{li2024playground}; (3) efficient generative models, such as SDXL-Turbo~\cite{sauer2023adversarial}, LCM-SDXL, LCM-SSD1B~\cite{luo2023latent}, LCM-PixArt~\cite{chen2024pixartdelta}, SDXL-Lightning~\cite{lin2024sdxl}; and (4) proprietary generative models, such as Dreamina~\cite{dreamina} and Midjourney v6.1~\cite{midjourney}. 
For each prompt, we randomly sample a subset of models for image generation, applying default parameters to ensure the quality of the generated images.  
This process results in  40K image-text pairs being generated using 4K prompts, ensuring a diverse dataset for annotation.

\begin{figure*}[htbp]
    \begin{minipage}[t]{0.31\textwidth}
        \centering
        \vspace{0pt} 
        \includegraphics[width=\linewidth]{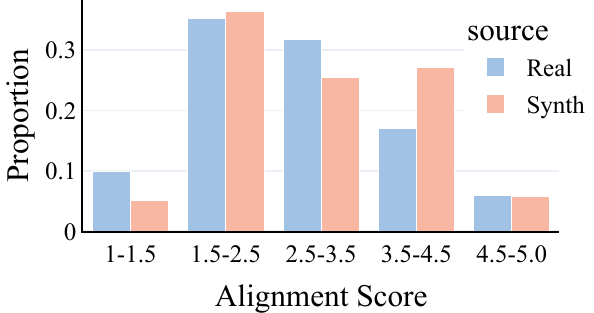}
    \end{minipage}
    \hspace{0.01\textwidth} 
    \begin{minipage}[t]{0.315\textwidth}
        \centering
        \vspace{0pt}
        \includegraphics[width=\linewidth]{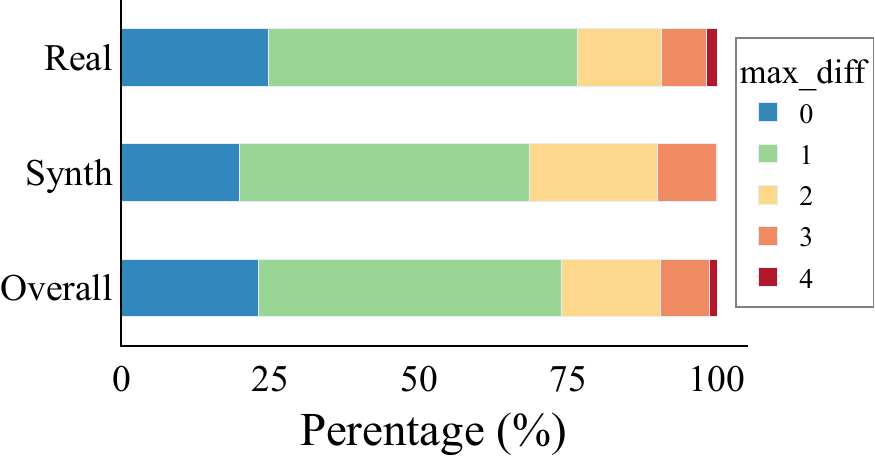}
    \end{minipage}
    \hspace{0.01\textwidth} 
    \begin{minipage}[t]{0.35\textwidth}
        \centering
        \vspace{0pt} 
        \includegraphics[width=\linewidth]{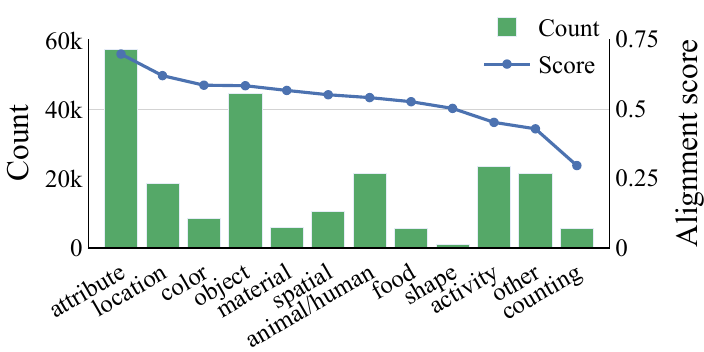}
    \end{minipage}
\end{figure*}

\begin{figure*}[!tbp]
    \vspace{-5mm}
    \begin{subfigure}[t]{0.31\textwidth}
        \centering
        \vspace{0pt} 
        \begin{minipage}{0.9\linewidth}
            \caption{}
            \label{fig:score}
        \end{minipage}
    \end{subfigure}
    \hspace{0.01\textwidth} 
    \begin{subfigure}[t]{0.315\textwidth}
        \centering
        \vspace{0pt} 
        \begin{minipage}{0.9\linewidth}
            \caption{}
            \label{fig:diff} 
        \end{minipage}
    \end{subfigure}
    \hspace{0.01\textwidth} 
    \begin{subfigure}[t]{0.35\textwidth}
        \centering
        \vspace{0pt} 
        \begin{minipage}{0.9\linewidth}
        \caption{}
        \label{fig:ele} 
        \end{minipage}
    \end{subfigure}
    \vspace{-3mm}
    \caption{Statistical Charts. (a) Distribution of annotated alignment scores in real prompt samples and synthetic prompt samples. (b) Distribution of maximum score differences between different annotators for the same image-text pair. (c) Distribution of element counts and scores across different skills in fine-grained annotations.}
    \vspace{-3mm}
\end{figure*}

\subsection{Data Annotation}
In this section, we describe how we perform the data annotation. We first define the content and templates of the annotations and then detail the entire annotation process.

\noindent
\textbf{Annotation Content and Templates.}
The annotation includes image-text alignment, structural problems, and extra information (see Fig.~\ref{fig:benchmark}(b)). 
In terms of image-text alignment, the annotator first scores the alignment using a 5-point Likert scale, like TIFA~\cite{hu2023tifa}.
Then, for the fine-grained elements in the prompt, the annotator needs to label whether they are aligned with the image or not. 
Annotators also mark structural issues in the generated images, categorized into three main types: humans, animals, and objects. For human figures, structural issues are further subdivided by specific regions, such as the face, hands, and limbs.
To address potential inaccuracies in element splitting by GPT-4, we introduce a new splitting confidence label, enabling annotators to flag instances with incorrect splitting during annotation.
Additionally, because some prompts originate from real users and may contain unclear meanings, we add a label to indicate whether a prompt is meaningful.

\noindent
\textbf{Annotation Process.}
To improve the quality of annotation, our annotation process is divided into three stages.
\textit{(1) Pre-annotation:} We formulate straightforward annotation standards to train the annotators, using a small amount of data for pre-annotation. 
The pre-annotated data are then carefully reviewed. 
Based on the issues encountered during the annotation and review process, we refine the evaluation standards to ensure greater clarity and consistency.
\textit{(2) Formal annotation:} During formal annotation, each image-text pair is labeled by three annotators, with an additional annotator assigned for quality control.
Furthermore, the annotators will identify any NSFW content in the generated images and determine whether they should be discarded.
\textit{(3) Re-annotation:} For instances where alignment scores from the three annotators show significant discrepancies (range $\ge2$), we conduct re-annotation to reduce subjective bias.
Ultimately, each image-text pair is annotated by 3 to 6 annotators, and the average score is used as the final label.

\subsection{Data Statistics and Reliability Analysis}

We generate 40K images using 4K prompts based on 20 T2I models. 
For each image-text pair, we perform multiple annotations, including image-text alignment scores, element matching, and structural issues, achieving nearly 1M annotations.
We employ novel sampling and generation strategies to ensure a balanced and diverse set of prompts. 
Additionally, we implement multi-round annotation and revision processes to guarantee the reliability of the annotated data.

We calculate the image-text alignment scores from the annotations and analyze the consistency of scoring across annotators.
The histogram of the alignment scores is shown in Fig.~\ref{fig:score}. 
The statistical results suggest that the alignment scores are distributed across all ranges, with a higher concentration in the middle range. 
This distribution provides a sufficient number of both positive and negative samples, enabling a robust evaluation of the consistency between existing image alignment metrics and human preferences. It also facilitates the training of a scoring model aligned with human preferences.

To analyze the agreement of the scoring data among annotators, we calculate the maximum score difference for each image-text pair and plot it as a histogram in Fig.~\ref{fig:diff}. 
It can be seen that 75\%  samples show a score difference of less than 1 point. 
It is worth noting that for score differences of 2 or more, we obtain double annotations (from 3 to 6) by re-annotating, further reducing the inter-annotator disagreement.

For fine-grained annotation, we perform statistics on quantity and alignment scores of elements based on their respective categories.
It can be observed from Fig.~\ref{fig:ele} that the overall alignment scores for most categories are around 50\%, ensuring a balanced distribution of positive and negative samples in fine-grained annotation. 
Additionally, we observe that the images generated by AIGC models exhibit relatively poor consistency with the text in aspects of counting, shape, and activity.

\section{Methods for Alignment Evaluation}

In this section, we introduce two methods for evaluating image-text alignment in AIGC tasks.
The proposed \textbf{FGA-BLIP2} can achieve \textbf{F}ine-\textbf{G}rained \textbf{A}lignment evaluation through end-to-end training. The proposed \textbf{PN-VQA} can perform zero-shot evaluation using a \textbf{P}ositive-\textbf{N}egative VQA manner.

\subsection{End-to-End Scoring Model: FGA-BLIP2}
Most scoring models for image-text alignment tasks are trained using reward model objectives, which learn human preferences between different images generated for the same prompt. 
Since alignment scores are not available during training, these methods often fail to produce values that accurately reflect the degree of image-text alignment.
Benefiting from the extensive image-text alignment scores labeled in EvalMuse-40K, we can train an end-to-end mapping between image-text pairs and their corresponding scores.

\begin{figure}[t]
    \centering
    \includegraphics[width=1\linewidth]{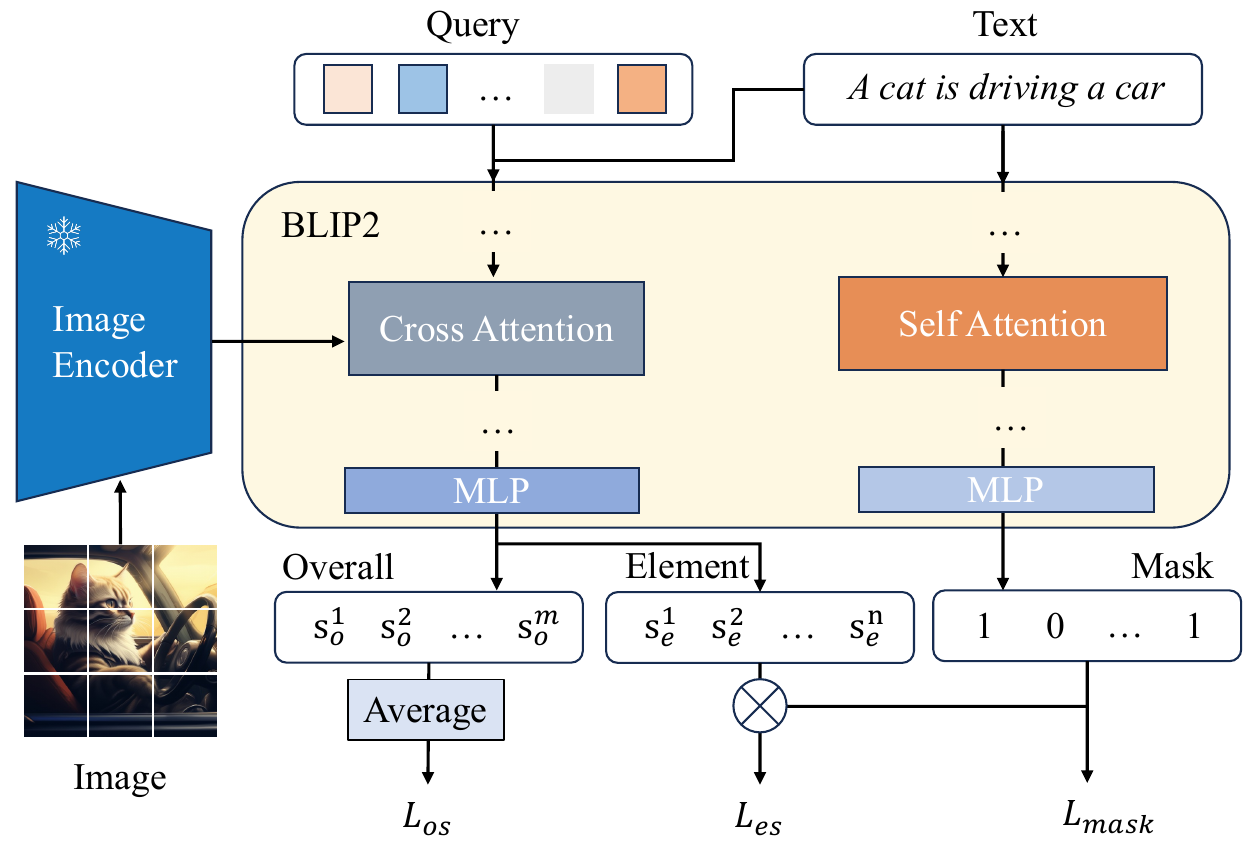}
    \caption{Structure of FGA-BLIP2. $m$ and $n$ denote query and text token lengths, respectively; $s_o$  and $s_e$ represent overall and element scores, respectively; $L_{\text{os}}$ represents the overall alignment score loss, $L_{\text{es}}$ is the element alignment score loss, and $L_{\text{mask}}$ is the loss for predicting valid elements. FGA-BLIP2 jointly optimizes these losses to achieve fine-grained evaluation.
    }
    \label{fig:blip2}
    \vspace{-2mm}
\end{figure}

Additionally, FGA-BLIP2  achieves fine-grained alignment evaluation by jointly training the overall and elemental alignment scores (see Fig.~\ref{fig:blip2}).
We adopt the setting of Image-Text Matching (ITM) in BLIP2~\cite{li2023blip}, where the query and text after embedding are concatenated and then processed through cross-attention with the image. This yields query embedding and text embedding outputs.
The final alignment score is obtained by a two-class linear classifier, where the query part is averaged to produce the overall alignment score, while the text part at each corresponding position provides the element-specific alignment scores. 
Since not all tokens in the text are highly relevant to the alignment task, we introduce an additional operation to predict valid tokens. 
The text is first fed to a self-attention layer to extract text features, which then are passed to an MLP to predict a mask that represents the validity of each text token.

We observe that some prompts in the dataset are either too simple or overly complex, resulting in minimal differences in alignment scores across images generated by different models. 
Such data may lead the model to focus more on prompt complexity than on the actual alignment level of the image-text pairs during training.
We therefore design a variance-weighted optimization strategy for the
image-text alignment task. 
We calculate the variance $\sigma(p)$ of the alignment scores across different images generated using the same prompt and use this to adjust the loss weights of different prompts during training.

The final loss objective function is as follows:
\begin{equation}
    L_{total} = e^{\sigma(p)} \cdot (L_{os} + \lambda L_{es} + \eta L_{mask}),
\end{equation}
where weighting parameters are set to $\lambda=0.1$ and $\eta = 0.1$. $L_{os}$ denotes the L1 loss between the predicted overall alignment score and human annotation. $L_{es}$ denotes the L1 loss between the predicted element score and human fine-grained annotation.  $L_{mask}$ denotes the L1 loss between the predicted valid text token and the real elements. 
When using variance to weight the loss function, a larger $\sigma(p)$ for an image-text pair results in a higher loss. This approach encourages the model to focus more on samples with greater differences in image-text matching scores under the same prompt, helping the model better understand and evaluate the alignment level between image and text.

\subsection{Fine-Grained Zero-Shot Alignment: PN-VQA}
Given the good performance of MLLMs, we follow the previous methods~\cite{hu2023tifa,yarom2024you} of using question answering for fine-grained evaluation. 
In comparison to the previous methods, we only generate simple yes/no judgment questions to evaluate whether the elements in the text are accurately represented in the generated image. By using straightforward yes/no questions, we effectively capture the VQA model's probability of outputting correct or incorrect answers.

Additionally, we utilize MLLMs to perform VQA tasks from both positive and negative perspectives. The specific question template is presented as follows:
\textit{``This image is generated from \{prompt\}. Is the answer to \{question\} in this image \{a\}?"}, where the answer $a \in \{yes,no\}$.
Then, we substitute the correct $a_T$ and the incorrect $a_F$ into this question template, and perform the VQA task to obtain the probabilities $P_T$ and $P_F$ that the model outputs a positive answer.
The formula for calculating the probability $P$ of an affirmative response from MLLMs is as follows:
\begin{equation}
    P = \frac{exp(L(I,Q,``Yes"))}{\sum_{A_{i} \in \{Yes,No\}}exp(L(I,Q,A_{i}))},
    \label{eq:S_e}
\end{equation}
where $L(I,Q,A)$ represents the logit of the output $A$ from the MLLMs when given the image $I$ and the question $Q$.

We then average to obtain the fine-grained alignment score as $S_e = (P_T + 1 - P_F) / 2$. 
This positive-negative questioning method enables MLLMs to perform a more robust fine-grained evaluation of generated images.
Additionally, we found that incorporating the original prompt into the question enables MLLMs to more effectively assess the alignment of elements with the image.

\begin{table*}[htbp]
  \centering
    \caption{Quantitative comparison between our methods and the state-of-the-art methods which only use image-text pair to output overall alignment score on multiple benchmarks. Here, `var' refers to the variance optimization strategy,  `os' represents the overall alignment score output by FGA-BLIP2, and `es\_avg' is the average of the element scores output by FGA-BLIP2. }
  \label{tab:method}%
  \resizebox{0.9\textwidth}{!}{
    \begin{tabular}{l|cc|cc|cc|cc}
    \hline
    \multirow{2}[1]{*}{Method} & \multicolumn{2}{c|}{\textbf{EvalMuse-40K (ours)}} & \multicolumn{2}{c|}{GenAI-Bench~\cite{li2024genai}} & \multicolumn{2}{c|}{TIFA~\cite{hu2023tifa}} & \multicolumn{2}{c}{RichHF~\cite{liang2024rich}} \\
\cline{2-9}          & SRCC  & PLCC  & SRCC  & PLCC  & SRCC  & PLCC  & SRCC  & PLCC \\
    \hline
    CLIPScore~\cite{hessel2021clipscore} & 0.2993 & 0.2933 & 0.1676 & 0.203 & 0.3003 & 0.3086 & 0.057 & 0.3024 \\
    BLIPv2Score~\cite{hessel2021clipscore} & 0.3583 & 0.3348 & 0.2734 & 0.2979 & 0.4287 & 0.4543 & 0.1425 & 0.3105 \\
    ImageReward~\cite{xu2024imagereward} & 0.4655 & 0.4585 & 0.34  & 0.3786 & 0.6211 & 0.6336 & 0.2747 & 0.3291 \\
    PickScore~\cite{kirstain2023pick} & 0.4399 & 0.4328 & 0.3541 & 0.3631 & 0.4279 & 0.4342 & 0.3916 & 0.4133 \\
    HPSv2~\cite{wu2023human} & 0.3745 & 0.3657 & 0.1371 & 0.1693 & 0.3647 & 0.3804 & 0.1871 & 0.2577 \\
    VQAScore~\cite{li2024evaluating} & 0.4877 & 0.4841 & 0.5534 & 0.5175 & 0.6951 & 0.6585 & 0.4826 & 0.4094 \\
    \hline
    FGA-BLIP2 (w/o var, os) & 0.7708 & 0.7698 & 0.5548 & 0.5589 & 0.7548 & 0.741 & 0.5073 & 0.5384 \\
 FGA-BLIP2 (es\_avg) & 0.6809 & 0.6867 & 0.5206 & 0.5259 & 0.7419 & 0.736 & 0.3413 &0.3096 \\
    FGA-BLIP2 (os) & \textbf{0.7742} & \textbf{0.7722} & 0.5637 & 0.5673 & 0.7604 & 0.7442 & \textbf{0.5123} & \textbf{0.5455} \\
    FGA-BLIP2 (os+es\_avg) & 0.7723 & 0.7716 & \textbf{0.5638} & \textbf{0.5684} & \textbf{0.7657} & \textbf{0.7508} & 0.4576 & 0.4967 \\
    \hline
    \end{tabular}%
  }
\end{table*}%
\section{Experiments} \label{5}

\begin{table*}[htbp]
  \centering
    \caption{Quantitative comparison between our methods and the state-of-the-art methods for fine-grained evaluation on EvalMuse-40K. Here, we report the correlation of the method on overall alignment scores and its accuracy on fine-grained alignment. Element-GT refers to the manually annotated fine-grained scores. `es' represents the element alignment
score output by FGA-BLIP2. $^*$ indicates that this method uses a fixed-step
(0.01) search to select the optimal threshold for binary classification, aiming to maximize overall accuracy.}
  \label{tab:vqa}%
  \resizebox{0.8\textwidth}{!}{
    \begin{tabular}{l|c|cc|cc|cc}
    \hline
    \multirow{2}[1]{*}{Method} & \multirow{2}[1]{*}{MLLMs} & \multicolumn{2}{c|}{Overall }& \multicolumn{2}{c|}{Real}&\multicolumn{2}{c}{Synth}\\
\cline{3-8}          &       & SRCC  & Acc (\%)& SRCC& Acc (\%)& SRCC& Acc (\%)\\
    \hline
    \multirow{3}{*}{TIFA~\cite{hu2023tifa}} & LLaVA1.6~\cite{liu2024llava} & 0.2937 & 62.1& 0.2348& 62.6& 0.4099& 60.6\\
          & mPLUG-Owl3~\cite{ye2024mplug} & 0.4303 & 64.5& 0.3890& 64.5& 0.5197& 64.4\\
          & Qwen2-VL~\cite{wang2024qwen2} & 0.4145 & 64.5& 0.3701& 64.4& 0.5049& 64.7\\
    \hline
    \multirow{3}{*}{VQ2$^*$~\cite{yarom2024you}} & LLaVA1.6~\cite{liu2024llava} & 0.4749 & 67.5& 0.4499& 67.2& 0.5314& 68.4\\
         & mPLUG-Owl3~\cite{ye2024mplug} & 0.5004 & 66.4& 0.4458& 65.8& 0.6145& 68.0\\
         & Qwen2-VL~\cite{wang2024qwen2} & 0.5415 & 67.9& 0.4893& \underline{67.3}& 0.6653& 67.0\\
    \hline
    \multirow{3}{*}{PN-VQA$^*$ (ours)} & LLaVA1.6~\cite{liu2024llava} & 0.4765 & 66.1& 0.4347& 65.5& 0.5486& 67.7\\
          & mPLUG-Owl3~\cite{ye2024mplug} & 0.5246 & 67.6& 0.5044& 67.1& 0.6032& 69.0\\
          & Qwen2-VL~\cite{wang2024qwen2} & \underline{0.5748} & \underline{68.2}& \underline{0.5315}& 67.0& \underline{0.6946}& \underline{71.9}\\
    \hline
    FGA-BLIP2 (es, ours) & BLIP2~\cite{li2023blip} & \textbf{0.6800} & \textbf{76.8}& \textbf{0.6298}& \textbf{75.9}& \textbf{0.7690}& \textbf{79.6}\\
    \hline
    Element-GT &  -     & 0.7273 & - & 0.6891 &    -   &   0.7839    &    -   \\
    \hline
    \end{tabular}%
    }
\end{table*}%
\subsection{Experimental Setup}
We use one-quarter of the samples from EvalMuse-40K as test set, ensuring no overlap in prompts between the training and test sets. 
The test set includes 500 real prompts and 500 synthetic prompts. 
We train FGA-BLIP2 on the training set and then test existing evaluation methods along with our FGA-BLIP2 and PN-VQA on the test set.
Additionally, we use GenAI-Bench~\cite{li2024genai}, TIFA~\cite{hu2023tifa}, and RichHF~\cite{liang2024rich} datasets to validate the generalization capability of our fine-tuned model FGA-BLIP2.

\noindent \textbf{Training Settings.} We use the BLIP-2~\cite{li2023blip} model fine-tuned on the COCO~\cite{lin2014microsoft} for the retrieval task as initialization weights. We follow BLIP-2's training setup to set the maximum learning rate to $1e^{-5}$ and the minimum learning rate to $0$ and train 5 epochs on an A100 GPU.

\noindent \textbf{Evaluation Settings.} 
For overall alignment scores, we compare FGA-BLIP2 with the state-of-the-art models~\cite{hessel2021clipscore,xu2024imagereward,kirstain2023pick,wu2023human,li2024evaluating} and report the Spearman Rank Correlation Coefficient (SRCC) and Pearson Linear Correlation Coefficient (PLCC) to measure the correlation between model predictions and human annotations.
For fine-grained evaluation, we compare PN-VQA with TIFA~\cite{hu2023tifa} and VQ2~\cite{yarom2024you}, and use several advanced MLLMs~\cite{liu2024llava,ye2024mplug,wang2024qwen2} for the VQA task.
On one hand, we average the fine-grained scores of the image-text pairs and compare them with the overall alignment scores from human annotations. 
On the other hand, we conduct fine-grained evaluation by reporting the accuracy of the method's element-wise predictions.

\begin{figure*}
    \centering
    \begin{minipage}{0.47\textwidth}
        \centering
        \includegraphics[width=\linewidth]{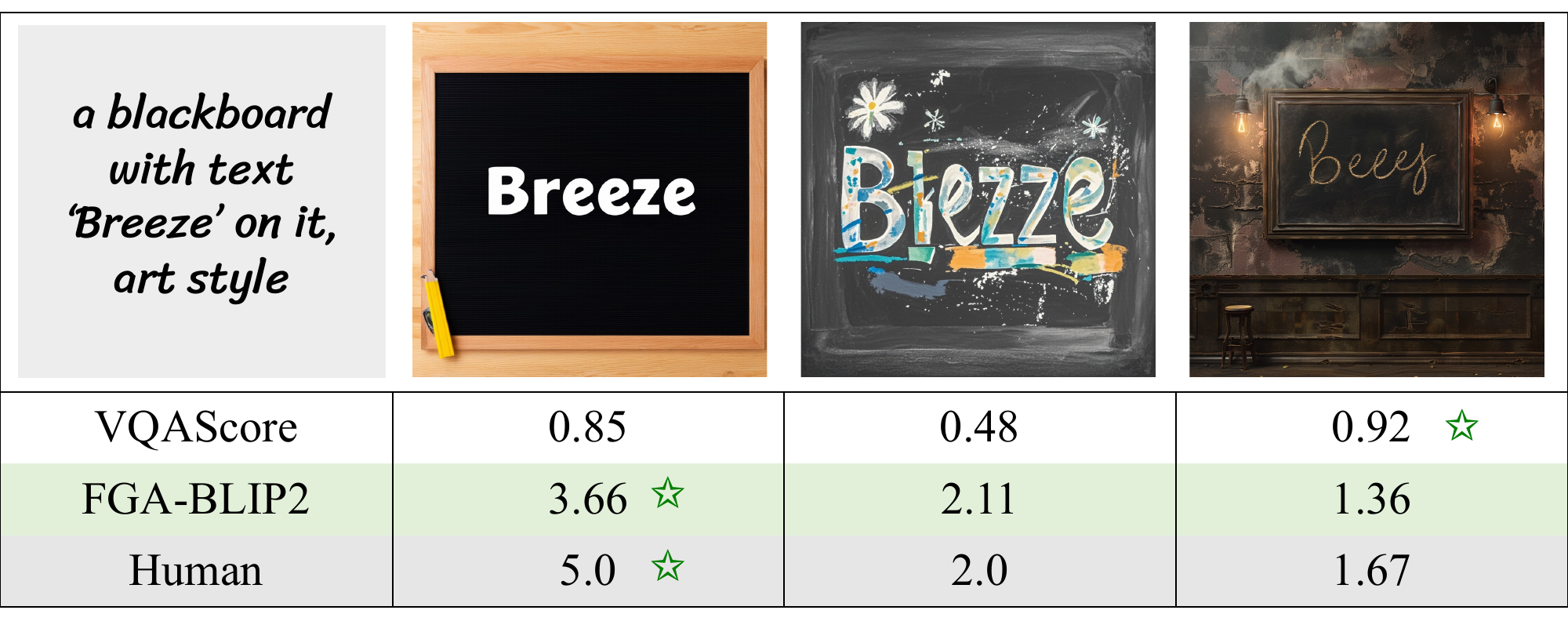}
    \end{minipage}
    \begin{minipage}{0.04\textwidth}
        \centering
    \end{minipage}
    \begin{minipage}{0.47\textwidth}
        \centering
        \includegraphics[width=\linewidth]{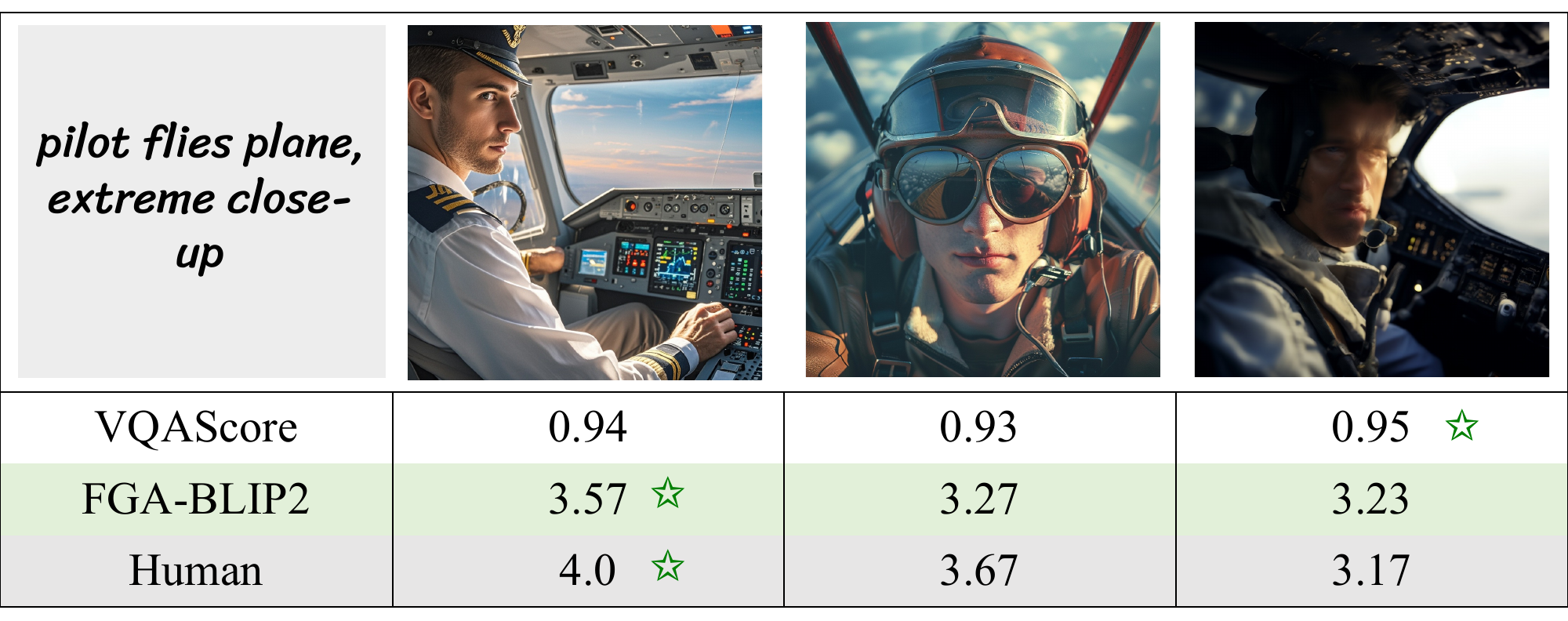}
    \end{minipage}
    \vspace{-2mm}
    \caption{Visualization of overall alignment scores from different methods. \textcolor{darkgreen}{\ding{73}} represents the image-text pairs with the highest alignment scores within the same method. ``Human" refers to the average alignment scores from multiple annotators. In comparison to VQAScore~\cite{li2024evaluating}, our FGA-BLIP2 aligns more closely with human annotations, providing a more accurate evaluation of image-text alignment.}
    \label{fig:vis_score}
    \vspace{-2mm}
\end{figure*}
\begin{figure*}
    \centering
    \begin{minipage}{0.47\textwidth}
        \centering
        \includegraphics[width=\linewidth]{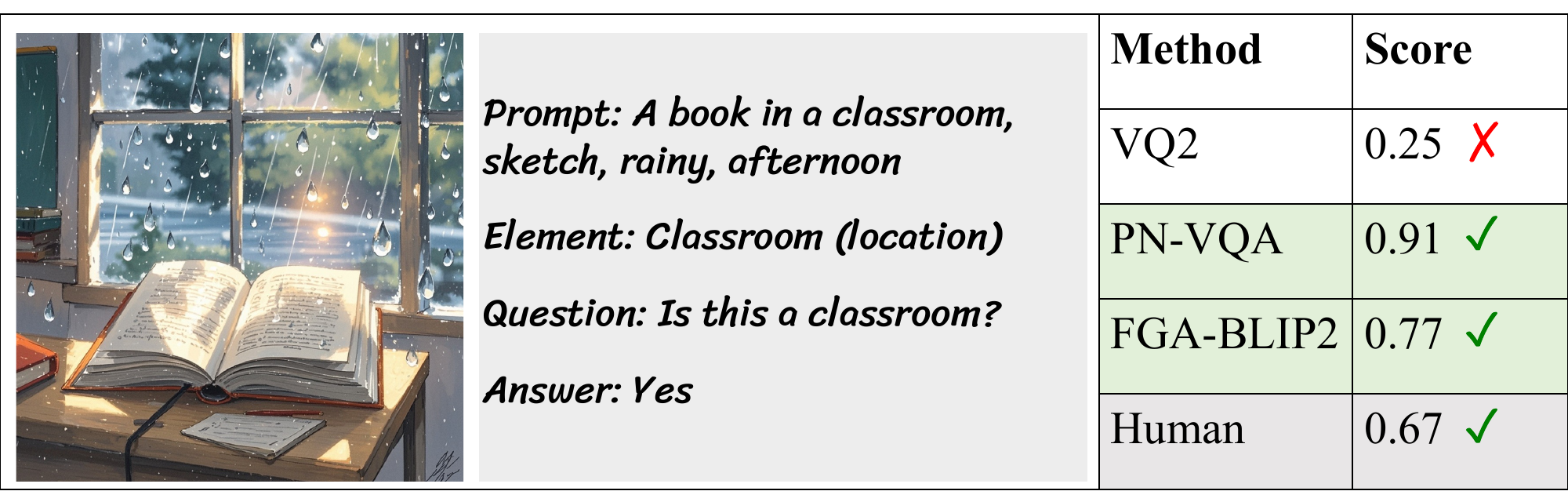}
    \end{minipage}
    \begin{minipage}{0.04\textwidth}
        \centering
    \end{minipage}
    \begin{minipage}{0.47\textwidth}
        \centering
        \includegraphics[width=\linewidth]{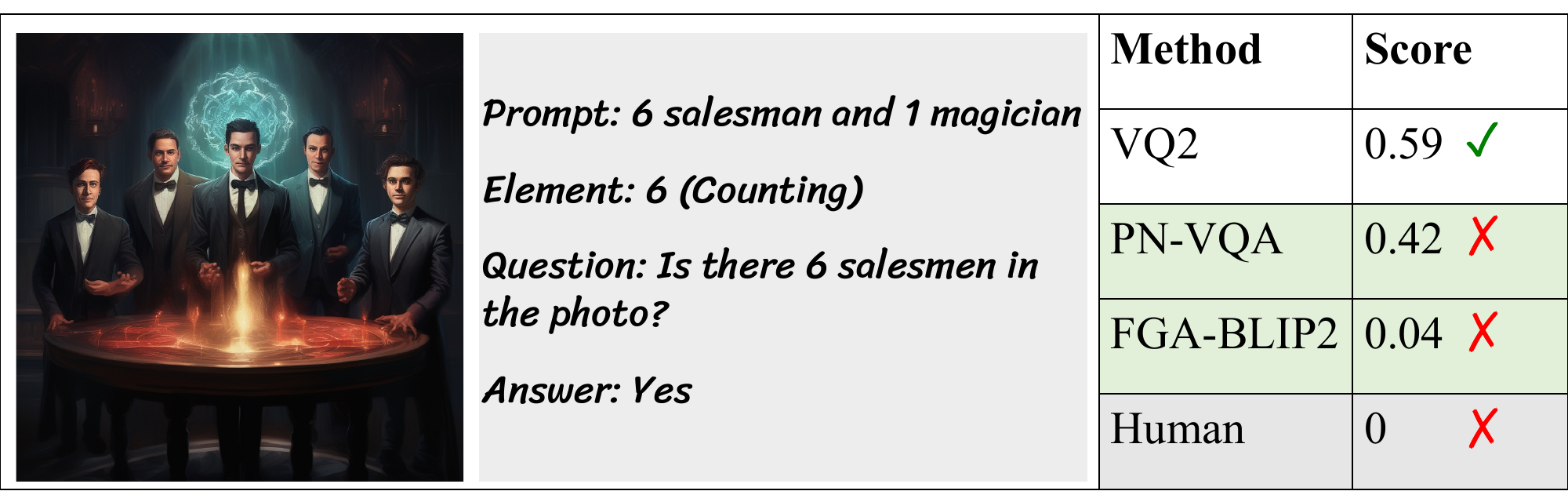}
    \end{minipage}
    \vspace{-2mm}
    \caption{Visualization of fine-grained alignment scores from different methods. In each figure, the question and answer are generated based on the element in the prompt and the score represents the probability of the element in the prompt matching the image (\cmark indicates a match, and \xmark indicates a mismatch). Here, both VQ2~\cite{yarom2024you} and PN-VQA use Qwen2-VL~\cite{wang2024qwen2} for VQA task. ``Human" refers to the average fine-grained scores from multiple annotators. It can be seen that our FGA-BLIP2 and PN-VQA yield results consistent with human annotations.}
    \label{fig:vis_ele}
    \vspace{-2mm}
    
\end{figure*}
\subsection{Results on Automated Metrics}

\noindent \textbf{Quantitative Analysis.}
We evaluate our proposed metrics from two aspects: overall alignment evaluation and fine-grained evaluation.

\noindent \textit{Overall Alignment.}
In Tab.~\ref{tab:method}, we report the results of FGA-BLIP2 and the state-of-the-art methods on multiple benchmarks.
The methods used here can directly output alignment scores from image-text pairs without extra operations like VQA. 
FGA-BLIP2 performs excellently not only on the EvalMuse-40K test set but also exhibits higher correlation with human annotations on GenAI-Bench~\cite{li2024genai}, TIFA800~\cite{hu2023tifa}, and the alignment score sections of RichHF~\cite{liang2024rich} compared to other methods.

\noindent \textit{Fine-Grained Evaluation.} 
In Tab.~\ref{tab:vqa}, we report the comparison results of our FGA-BLIP2 and PN-VQA with other VQA methods such as TIFA~\cite{hu2023tifa} and VQ2~\cite{yarom2024you}. 
We use SRCC to measure the correlation between averaged fine-grained element scores and image-text alignment scores, and report the accuracy of fine-grained evaluation. 
It can be seen that FGA-BLIP2 achieves the best performance, and PN-VQA also performs well compared to other VQA-based methods.
In addition, Qwen2~\cite{wang2024qwen2} yields a superior correlation with human annotations than other VQA models.

For different elemental skills in fine-graining such as counting and color, we also compare multiple fine-graining evaluation methods, and both FGA-BLIP2 and PN-VQA achieve good performance. Meanwhile, we select 200 representative prompts and use our FGA-BLIP2 to evaluate the alignment of image-text pairs generated by over 20 T2I models. We found that current proprietary models such as Midjourney~\cite{midjourney} and Dreamina~\cite{dreamina} generally performed better. Additionally, we evaluate MLLMs' ability to detect structural problems using EvalMuse-40K's structural labels. Due to space constraints, the comparison results for all element skills, the ranking of T2I models' alignment ability by FGA-BLIP2, and the exploration of different MLLMs' structural problem detection abilities are provided in the supplementary material.

\noindent \textbf{Qualitative Analysis.}
In Fig.~\ref{fig:vis_score}, we present the prediction scores of FGA-BLIP2 and VQAScore~\cite{li2024evaluating} across multiple image-text pairs. It can be seen that FGA-BLIP2 produces scores that are more consistent with human annotations. 
For fine-grained evaluation, as shown in Fig.~\ref{fig:vis_ele}, our proposed methods are able to more accurately determine whether the elements in the prompt are represented in the generated image.

\subsection{Ablation Study}
The ablation study for FGnA-BLIP2 is shown in Tab.~\ref{tab:method}.
When evaluating using the overall alignment scores (os) output by FGA-BLIP2, our model achieves better correlation across four datasets compared to the method without the variance optimization strategy (w/o var).
We also report the fine-grained element scores (es) output by FGA-BLIP2 and take the average (avg) as the image-text alignment score for evaluating the model's performance on the datasets.
Combining and weighting os and es\_avg as the final score leads to improved performance on GenAI-Bench and TIFA.

The ablation study for PN-VQA is shown in Tab.~\ref{tab:ablation-vqa}. 
The results show that incorporating the prompt used for generating the image into the question, and performing VQA in both positive and negative question manners separately, significantly improves the correlation between the VQA method and human annotations.
\begin{table}[htbp]
  \centering
      \caption{Ablation study for PN-VQA using Qwen2-VL~\cite{wang2024qwen2}. The table shows the correlation of overall alignment scores with human annotations.}
  \label{tab:ablation-vqa}%
    \begin{tabular}{lccc}
    \hline
    Setting & SRCC  & KRCC  & PLCC \\
    \hline
    PN-VQA (w/o prompt) & 0.5576 & 0.4058 & 0.5583 \\
    PN-VQA (positive) & 0.5406 & 0.3877 & 0.5271 \\
    PN-VQA (negative) & 0.5556 & 0.4036 & 0.5534 \\
    PN-VQA & \textbf{0.5748} & \textbf{0.4170} & \textbf{0.5694} \\
    \hline
    \end{tabular}%

\end{table}%

\section{Conclusion}

In this work, we contribute EvalMuse-40K, which contains a large number of manually annotated alignment scores and fine-grained element scores, enabling a comprehensive evaluation of the correlation between automated metrics and human judgments in image-text alignment-related tasks.
We also propose two new evaluation methods: FGA-BLIP2, which fine-tunes a vision-language model to output token-level alignment scores, and PN-VQA, which improves the ability of the VQA model in fine-grained perception by positive-negative questioning.
These methods improve the correlation between current image-text alignment metrics and human annotations, enabling better evaluation of image-text alignment in T2I models.

{
    \small
    \bibliographystyle{ieeenat_fullname}
    \bibliography{main}
}

\clearpage
\setcounter{page}{1}
\maketitlesupplementary

\setcounter{section}{6} %
\section{Overview}
In this supplementary material, we provide \textbf{data collection details}, \textbf{additional experiment results}, and \textbf{T2I model alignment evaluation}. 
In data collection details,  we detail the classification and sampling of real user prompts in Sections~\ref{sec:real} and \ref{milp}, ensuring diversity and balance among the real prompts. We then describe the specific method for generating synthetic prompts in Section~\ref{sec:synth} and detail the element splitting and question generation processes in Sections~\ref{sec:ele} and ~\ref{sec:quest}. 
Regarding additional experimental results, we report the performance of FGA-BLIP2 and PN-VQA on fine-grained specific skills (see Tab.~\ref{tab:acc_fine}) and provide additional visualizations in Sections~\ref{sec:skill_eval} and ~\ref{sec:more_vis}. We evaluate the structure-aware capabilities of MLLMs using the structural labels in EvalMuse-40K in Section~\ref{sec:struct}.
Finally, in T2I model alignment evaluation, we employ FGA-BLIP2 to assess more than 20 widely used T2I models. The evaluation processes and results are presented in Section~\ref{sec:t2i_eval} and Tab.~\ref{tab:t2i}.

\begin{table*}[!tbp]
  \centering
    \caption{Samples of each category in the different dimensions of real prompts.}
    \begin{tabular}{ccl}
    \toprule
    \textbf{Dimension} & \textbf{Type} & \textbf{Prompt} \\
    \midrule
    \multirow{13}[2]{*}{Subject} & Artifacts & sharp design spaceship illustrations, sketches \\
          & World Knowledge & The Community (2009) movie, poster \\
          & People & cute anime girl sleeping, high quality, award winning, digital art \\
          & Outdoor Scenes & highly detailed photo of road going between two mountains, photo realistic \\
          & Illustrations & Illustration of a baby daikon radish in a tutu walking a dog \\
          & Vehicles & A slit-scan photo of a car moving fast by Ted Kinsman \\
          &  Food \& Beverage & a purple french baguette \\
          & Arts  & ink painting of a single cowboy in style of Sin City by Frank Miller \\
          & Abstract & Trapped in a dream \\
          & Produce \& Plants & inside the giant glass geodesic dome full of forest lush vegetables \\
          & Indoor Scenes & photo of a small modern bathroom, isometric perspective, bird's eye view \\
          & Animals & a giant bunny walking in the street \\
          & Idioms & A storm in a teacup \\
    \midrule
    \multirow{6}[2]{*}{Logic} & Position Relationship & a beaver, sitting at a computer, playing video games, comic sketch  \\
          & Number & a beautiful oil painting of two ducks in a pond \\
          & Color & black resin flowers, yellow earth, red sky, impressionism \\
          & Writing \& Symbols & a cake saying happy birthday \\
          & Perspective & supermarket aisles, fisheye lens, color, fluorescent lighting \\
          & Anti-reality & astronaut carries little horse on his spine \\
    \midrule
    \multirow{5}[2]{*}{Style} & Material & a house built entirely from huge diamonds \\
          & Genre & race cars, style of red line anime movie, centered \\
          & Design & a detailed illustration of a cupcake made of unicorns \\
          & Photography \& Cinema & thousands of oranges, fruit, product photo \\
          & Artist \& Works & a sunset in the style of your name \\
    \bottomrule
    \end{tabular}%
  \label{tab:sample_real}%
\end{table*}%

\section{Data Collection Details}
\subsection{Real Prompt Categorization}\label{sec:real}
DiffusionDB~\cite{wang2022diffusiondb} contains a large and diverse set of real user-used prompts. Our goal is to select a representative subset of prompts that accurately reflects the overall distribution. To achieve this, we classify the prompts and use sampling to ensure balance across categories. After careful consideration, we choose four key dimensions for classification: subject categories, logical relationships, image styles, and BERT embeddings.

Specifically, \textbf{subject categories} are divided into the following types: Artifacts, World Knowledge, People, Outdoor Scenes, Illustrations, Vehicles, Food \& Beverage, Arts, Abstract, Produce \& Plants, Indoor Scenes, Animals, and Idioms. \textbf{Logical relationships} are divided into Position Relationship, Number, Color, Writing \& Symbols, Perspective, and Anti-reality. \textbf{Image  style} includes Material, Genre, Design, Photography \& Cinema, and Artist \& Works. \textbf{BERT embeddings} are used to represent the semantic information of the prompts.

The prompt collection process is as follows. We first randomly select 100K prompts from DiffusionDB as the source dataset. GPT-4~\cite{achiam2023gpt} is then utilized to label the three dimensions of subject category, logical relationship, and image style of the prompt representation, meanwhile allowing the prompt to have multiple labels in a given dimension. In Tab.~\ref{tab:sample_real}, we show examples of real prompts from various categories. The semantic dimensions represented by the prompt's BERT embeddings are then categorized into seven types using the K-Means clustering algorithm.

\subsection{Prompt Sample Strategy} \label{milp}
We use Mixed-Integer Linear Programming (MILP) to sample dataset, ensuring a balanced distribution across different dimensions.
We define $ M $ as the number of data dimensions, $ H_m $ as the number of categories in each dimension, $ K $ as the number of original data samples, and $ N $ as the number of samples to be obtained after sampling.
Let $ B = \{ B_m \}_{m=1}^M $ denote a set of binary matrices, with $ B^m \in \mathbb{Z}_{2}^{H_m \times K} $, where each binary element $ b_{ij} $ indicates whether the $ j $-th sample in the original data belongs to the $ i $-th category in dimension $ m $.
Let $D = \{D_m\}_{m=1}^M$, where $D_m \in \mathbb{R}^{H_m}$ represents the distribution ratio of each category in the $m$-th dimension of the sampled data.
Finally, we introduce a binary vector $ \mathbf{x} \in \mathbb{Z}_2^K $, where the coefficient $ x_i $ indicates whether the $ i $-th sample in the original dataset is selected during sampling. 
The objective of our sampling is to ensure that the sampled data conforms to the distribution $ D $.
The problem then can be formulated as the following minimization:
\begin{equation}
    \min_{x} \sum_{m=1}^{M}||B^m \mathbf{x} - ND_m||_1 \, \, s.t. \, ||\mathbf{x}||_1 = N,
\end{equation}
which involves selecting $N$ samples from the original data such that their distribution closely aligns with $D$.
The above minimization can be solved by using a set of auxiliary vectors $ \mathbf{Z} = \{ \mathbf{z}_m \}_{m=1}^{M} $ with $ \mathbf{z}_m \in \mathbb{R}^{H_m}_+ $, in order to handle the absolute values of the $ L_1 $ norm~\cite{vonikakis2016shaping}:
\begin{equation}
\resizebox{0.9\linewidth}{!}{$
\left\{\hspace{-2mm}
\begin{array}{l}
B^m\mathbf{x} - ND_m \leq \mathbf{z}_m \\
B^m\mathbf{x} - ND_m \geq -\mathbf{z}_m
\end{array}
\right.
\hspace{-4mm}
\xrightarrow{}
\left\{\hspace{-2mm}
\begin{array}{l}
 B^m\mathbf{x} - \mathbf{z}_m \leq ND_m  \\
-B^m\mathbf{x} - \mathbf{z}_m  \leq -ND_m
\end{array}
\right.
$}
\end{equation}
for each dimension $m$ and minimizing over $Z$. The final optimization can be expressed as MILP as:
\begin{equation}
    \min \mathbf{c}^{T}\Tilde{\mathbf{x}} \, \, s.t. \, A\Tilde{\mathbf{x}} \leq \mathbf{b}
\end{equation}
with $\mathbf{c} = [\mathbf{0}_K^T \,\, \mathbf{1}_{H_1}^T \dots \mathbf{1}_{H_M}^T]^T$
, $\Tilde{\mathbf{x}}=[\mathbf{x}^T \, \, \mathbf{z}_1^T \dots \mathbf{z}_M^T]^T$ and
\[
A = \left[ \begin{array}{cccc}
\mathbf{1}^T_K  &  \mathbf{0}^T_{H_1} & \dots & 0^T_{H_M} \\
 -\mathbf{1}^T_K  &  0^T_{H_1} & \dots & 0^T_{H_M} \\
\hline
B^{1} & -\mathbf{I}_{H_1}\\
\vdots & &\ddots\\
B^{M} & & & -\mathbf{I}_{H_M}\\
\hline
-B^{1} & -\mathbf{I}_{H_1}\\
\vdots & &\ddots\\
-B^{M} & & & -\mathbf{I}_{H_M}
\end{array} \right],
b = \left[ \begin{array}{c}
     N  \\
     -N \\
     \hline
     ND_1 \\
     \vdots \\
     ND_M \\
     \hline
     -ND_1 \\
     \vdots \\
     -ND_M
\end{array}
\right]
\]
where $A \in \mathbb{Z}^{(2+2\sum H_m) \times (K + \sum H_m)}$, $b \in \mathbb{R}^{2 + 2 \sum H_m}$, and $c \in \mathbb{Z}_2^{K + \sum H_m}$.
$\Tilde{\mathbf{x}}$ is also of size $K+ \sum H_m$ and contains both the integer and real optimization variables.

\begin{figure*}[!t]
    \includegraphics[width = \textwidth]{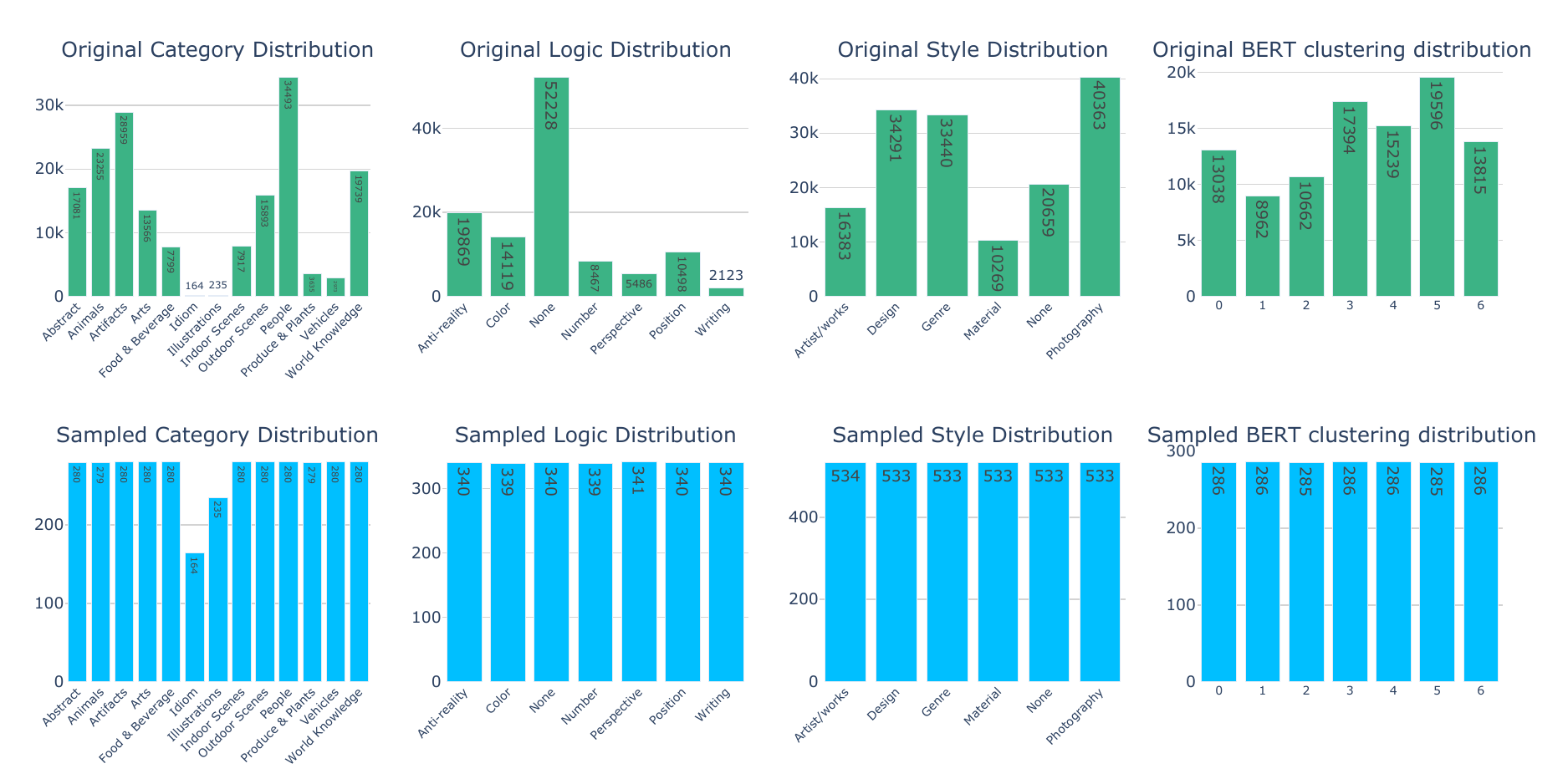}
    \caption{Distribution of data before (top) and after (bottom) sampling in four dimensions using MILP strategy.
    It can be observed that through MILP sampling, the distribution of prompts across various dimensions is more balanced.}
    \label{fig:sample}
\end{figure*}

In our sampling process, we set $K=100,000$ and $N=2,000$. To ensure a uniform distribution, the value of $D_m$ is set to $R_m / {H_m}$, where $R_m$ represents the average number of categories per prompt in dimension $m$, and $H_m$ denotes the total number of categories in dimension $m$.
Through the above sampling method, we ensure the diversity and balance of the sampled data. The distributions before and after sampling are shown in Fig.~\ref{fig:sample}.

\subsection{Sythetic Data Generation}\label{sec:synth}
We generate 2K prompts by using keyword combination and GPT-4. These prompts can be categorized into six different classes. 

\noindent 
\textbf{Naive Object Count.} 
We prepare a table that contains hundreds of object names. These objects can be classified into animal, human, object-household, object-electronic, object-cloth, object-others, object-vehicle, and food. 
To generate the prompt, we randomly select two numbers from 0-6 and two objects, then concatenate them. 
For example, given sampled number 1 and some, objects cat and dogs, the generated prompt should be ``1 cat and some dogs".

\noindent 
\textbf{Object Color Material.} 
We randomly select two objects and two colors from our keyword table. We use GPT-4 to generate a prompt following the fixed format which contains color, object, and material. 
The prompt is ``\textit{Generate a natural and reasonable prompt following the format [color] [adj-material][noun] and [color][adj-material][noun], given noun \{xx\} and color \{xx\}.}'' 
For example, given color red and green and the object cat and dog, the generated prompt can be ``a furry red cat and a green plastic dog".

\noindent 
\textbf{Object Location Weather Timeline Style.} 
We randomly select one object name, one location keyword, and one style keyword from our keyword table. 
The prompt is generated by GPT-4 by giving the following prompt: \textit{Generate a prompt following the format A/an [noun][adp] [location], [style], [weather], [time], given noun \{xx\}, location \{xx\} and style \{xx\}.} 
For example, given noun crew, location river, style 2D, the generated prompt should be ``A crew at the River, 2D style, sunny, afternoon".

\noindent 
\textbf{Attribute Object Activity Perspective View.} 
We randomly select one object name from the human and animal class, then use GPT-4 to generate a verb and a noun that correctly match the selected object. 
The prompt is shown as ``\textit{Generate a natural and reasonable prompt following the format [nounA] [verb][nounB], given nounA \{xx\}.}'' 
For example, given nounA: miner, the generated prompt should be ``miner digs ore". 
After that, we add one perspective view-related keyword to the end of the phrase. The keyword is randomly selected from our keyword lists. 
The final generated prompt should follow the format ``miner digs ore, panorama".

\noindent 
\textbf{Naive Writing Symbols.}
We collect a set of words and symbols.
We randomly select one element from the words-symbol set and select one word style keyword from [``Regular script font", ``Times New Roman font", ``Art style", ``Landscape orientation", ``Portrait orientation"]. 
Then we set the prompt as  \textit{``a blackboard with text \{xx\} on it, \{xx\} style"}. For example, given the word ``love" and style keyword ``Art style", the prompt should be ``a blackboard with text 'love' on it, Art Style".

\noindent 
\textbf{Attribute Spatial Composition.}
We randomly select two object names, one spatial keyword, and one attribute keyword, then use GPT-4 to generate the prompt that reflects the spatial and attribute relationship between these objects. 
The prompt is shown as ``\textit{Generate a natural and reasonable prompt following the format [nounA] [position] [noun B] and the [noun A] [adj] the [noun B], given noun \{xx\}. 
[position] denotes the positional relationship between objects and [adj] denotes the contrasting relationship between objects.}''
For example, given the noun boy and girl, the generated prompt may be: ``a boy is beside a girl, and the boy is taller than the girl".

\begin{figure*}[t]
\lstset{
    basicstyle=\ttfamily\footnotesize,  
    frame=single,                       
    breaklines=true,                     
    captionpos=b,                        
    tabsize=4,                           
    keepspaces=true,                     
    showspaces=false,                    
    showtabs=false,                      
    backgroundcolor=\color{gray!10},     
    aboveskip=10pt,                      
    belowskip=10pt,                      
    xleftmargin=15pt,                    
    xrightmargin=15pt,                   
    numbers=left,                        
    numberstyle=\tiny\color{gray},       
    stepnumber=1,                        
    numbersep=5pt,                       
    lineskip=0pt                         
}
\begin{lstlisting}[caption={LLM template for element splitting.},label={L:element}]
f"""
    Given an aigc prompt, extract the elements that are important for generating images.
    Classify each element into a type (object, human, animal, food, activity, attribute, counting, color, material, spatial, location, shape, other).
    Examples are as follows, where Elements is in json format.

    Prompt:A man posing for a selfie in a jacket and bow tie.
    Elements:["man (human)", "selfie (activity)", "jacket (object)", "bow tie (object)", "posing (activity)"]

    Prompt:A horse and several cows feed on hay.
    Elements:["horse (animal)", "cows (animal)", "hay (object)", "feed on (activity)", "several (counting)"]

    ...

    Prompt: {prompt}
    Elements: 
"""
\end{lstlisting}
\begin{lstlisting}[caption={LLM template for question generation.},label={L:question}]
f"""
    Given a prompt for image generation and one of its related elements, generate one easy Yes/No question to verify whether the element is represented in the image generated by the prompt.
    
    Description: A man posing for a selfie in a jacket and bow tie.
    Element: man (human):
    Q: Is this a man?
    Choices: yes, no
    A: yes
    
    Description: Several Face mask and 0 nun
    Element: 0 (Counting):
    Q: Are there any nuns in the photo?
    Choices: yes, no
    A: no
    
    ...
    
    Description: {caption}
    Element: {element}:
"""
\end{lstlisting}
\end{figure*}

\subsection{Element Splitting}\label{sec:ele}
For fine-grained annotation, we use GPT-4 to split the prompts into multiple elements and classify them according to the categories in TIFA~\cite{hu2023tifa}. The specific categories are as follows: object, human, animal, food, activity, attribute, counting, color, material, spatial, location, shape, and other. The template used is shown in Listing~\ref{L:element}.

\subsection{Question Generation}\label{sec:quest}
In question generation, we need to provide the prompt and the corresponding elements to be examined, and then use GPT-4 to generate simple binary questions, ensuring that the answer is either Yes or No. The question generation template used is shown in Listing~\ref{L:question}.

\section{More Experiments}
\setlength{\tabcolsep}{3pt} 
\begin{table*}[htbp]
  \centering
    \caption{Accuracy (\%) of fine-grained evaluation methods on different element categories. Align-Score represents the average fine-grained human annotation for each element category, indicating the degree of alignment with the image. \small{1},\small{2},\small{3} denote the use of Llava1.6~\cite{liu2024llava}, mPLug-Owl3~\cite{ye2024mplug}, Qwen2-VL~\cite{wang2024qwen2} for the VQA task, respectively. And a./h. is the abbreviation for animal/human. All the above methods use a fixed-step (0.01) search to select the optimal threshold for binary classification, aiming to maximize overall accuracy.}
    \begin{tabular}{lcccccccccccc}
    \toprule
          & overall & attribute & material & color & location & object & a./h. & food  & spatial & activity & shape & counting \\
    \midrule
    Align-Score &       & 0.706 & 0.663 & 0.636 & 0.609 & 0.564 & 0.519 & 0.517 & 0.507 & 0.452 & 0.438 & 0.283 \\
    \midrule
    TIFA$^1$  & 62.1  & \underline{74.7}  & 71.3  & 67.8  & 64.9  & 61.5  & 57.3  & 63.7  & 54.8  & 50.7  & 38.5  & 46.3 \\
    TIFA$^2$  & 64.5  & 69.6  & 70.0  & 69.2  & 71.1  & 66.9  & 65.0  & 64.8  & 55.1  & 56.4  & 48.6  & 57.3 \\
    TIFA$^3$  & 64.5  & 71.8  & 70.5  & \textbf{71.8} & 69.6  & 65.6  & 63.7  & 63.8  & 57.8  & 55.4  & 48.1  & 56.7 \\
    VQ2$^1$   & 67.5  & 69.6  & 68.3  & \underline{70.6}  & 72.5  & 68.7  & 67.6  & 69.3  & 59.1  & 65.0  & 51.0  & 65.2 \\
    VQ2$^2$   & 66.4  & 64.4  & 65.8  & 68.0  & 72.8  & 70.0  & 70.5  & 67.2  & 60.6  & 63.7  & \underline{53.8}  & 67.4 \\
    VQ2$^3$   & 67.9  & 65.9  & 70.3  & 70.1  & \underline{73.0}  & \underline{70.4}  & \underline{71.7}  & \underline{69.9}  & 61.4  & \underline{65.8}  & 52.4  & \underline{71.2} \\
    PN-VQA$^1$ & 66.1  &  71.9 & \underline{74.0}  & 68.3  & 70.8  & 66.4  & 68.1  & 67.0  & 59.8  & 57.5  & 53.4  & 59.0 \\
    PN-VQA$^2$ & 67.6  & 72.4  & 69.2  & 68.3  & 71.5  & 67.9  & 69.2  & 64.3  & 61.9  & 60.7  & 44.2  & 64.8 \\
    PN-VQA$^3$ & \underline{68.2}  & 70.7  & 71.4  & 69.8  & 71.8  & 68.8  & 70.7  & 66.6  & \underline{63.7}  & 62.3  & 49.5  & 66.0 \\
    FGA-BLIP2 & \textbf{76.8} & \textbf{80.8} & \textbf{80.1} & 68.2  & \textbf{76.2} & \textbf{73.8} & \textbf{75.5} & \textbf{72.9} & \textbf{72.8} & \textbf{76.3} & \textbf{67.8} & \textbf{79.7} \\
    \bottomrule
    \end{tabular}%

  \label{tab:acc_fine}%
\end{table*}%

\begin{figure*}
    \begin{minipage}{0.5\textwidth}
        \centering
        \includegraphics[width=\linewidth]{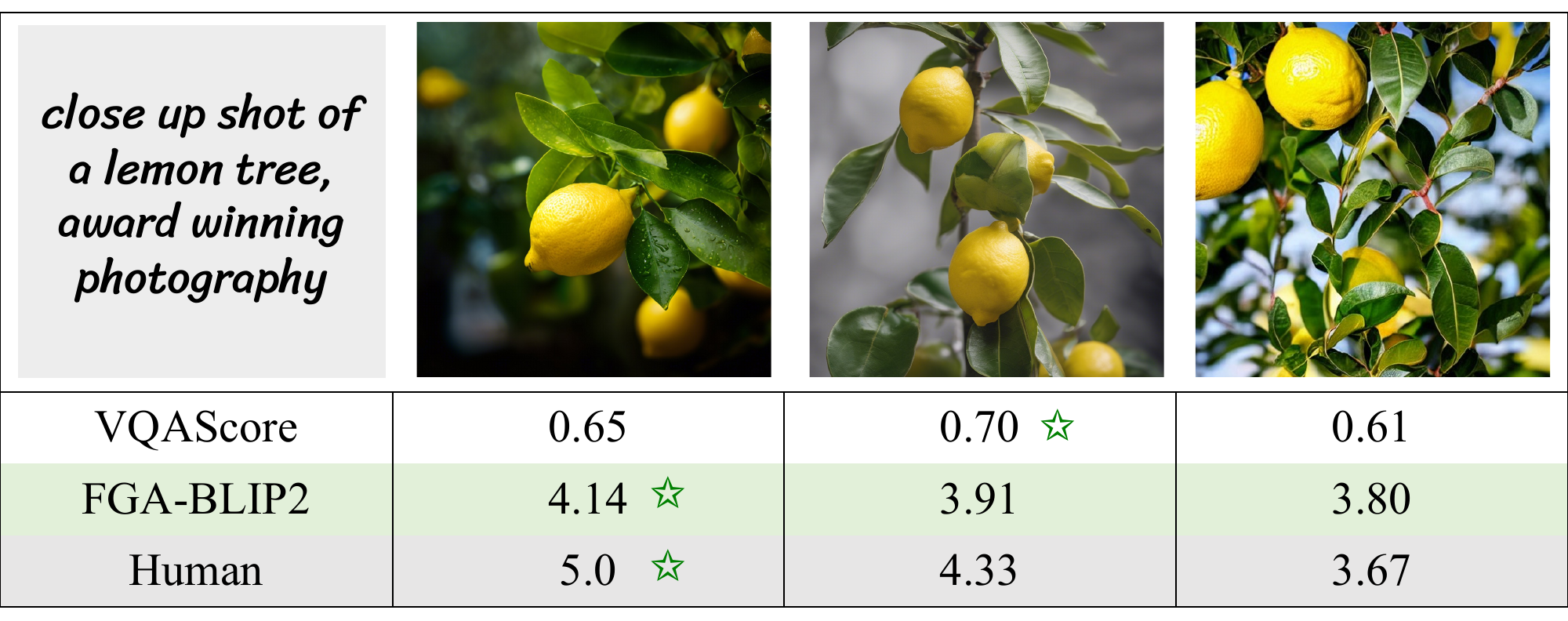}
    \end{minipage}
    \begin{minipage}{0.5\textwidth}
        \centering
        \includegraphics[width=\linewidth]{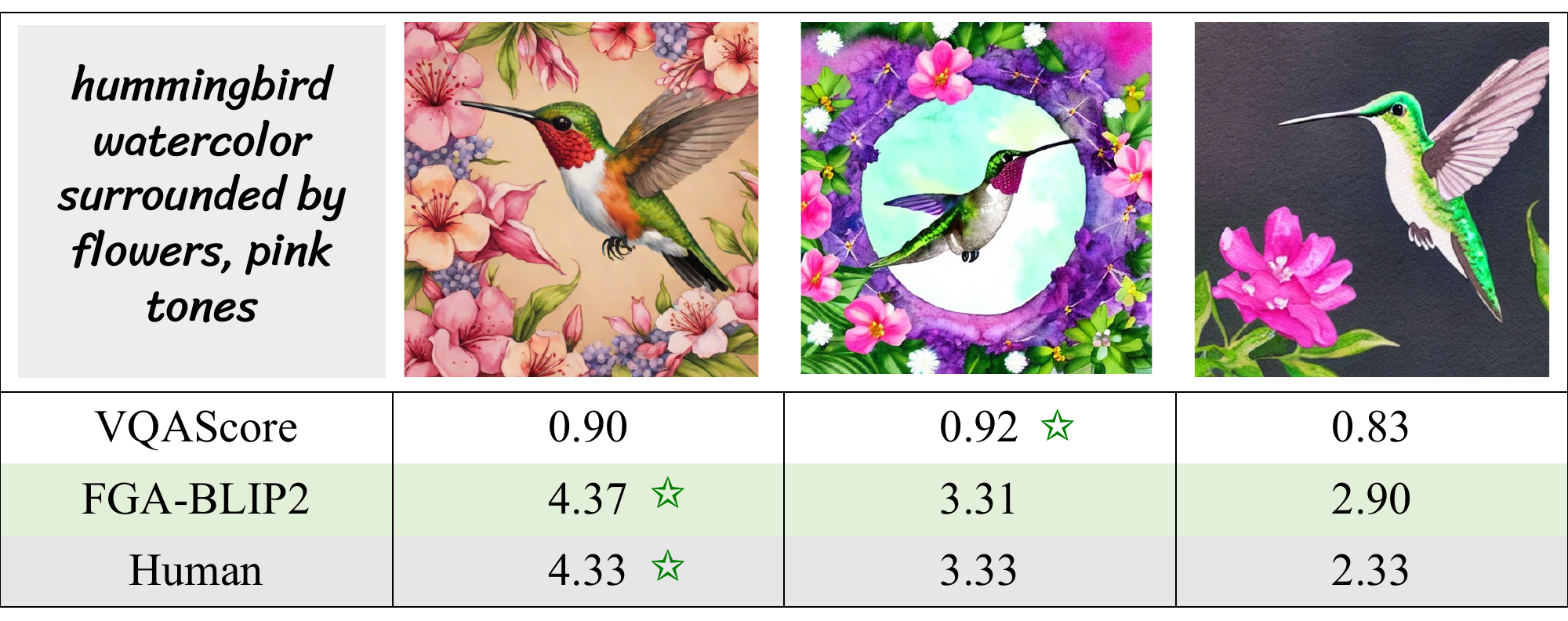}
    \end{minipage}
    \vspace{-2mm}
    \caption{Visualization of overall alignment scores from different methods. \textcolor{darkgreen}{\ding{73}} represents the image-text pairs with the highest alignment scores within the same method. ``Human" refers to the average alignment scores from multiple annotators. In comparison to VQAScore~\cite{li2024evaluating}, our FGA-BLIP2 aligns more closely with human annotations, providing a more accurate evaluation of image-text alignment.}
    \label{fig:vis_score_supp}
    \vspace{-2mm}
\end{figure*}
\begin{figure*}[!htbp]
    \begin{minipage}{0.5\textwidth}
        \centering
        \includegraphics[width=\linewidth]{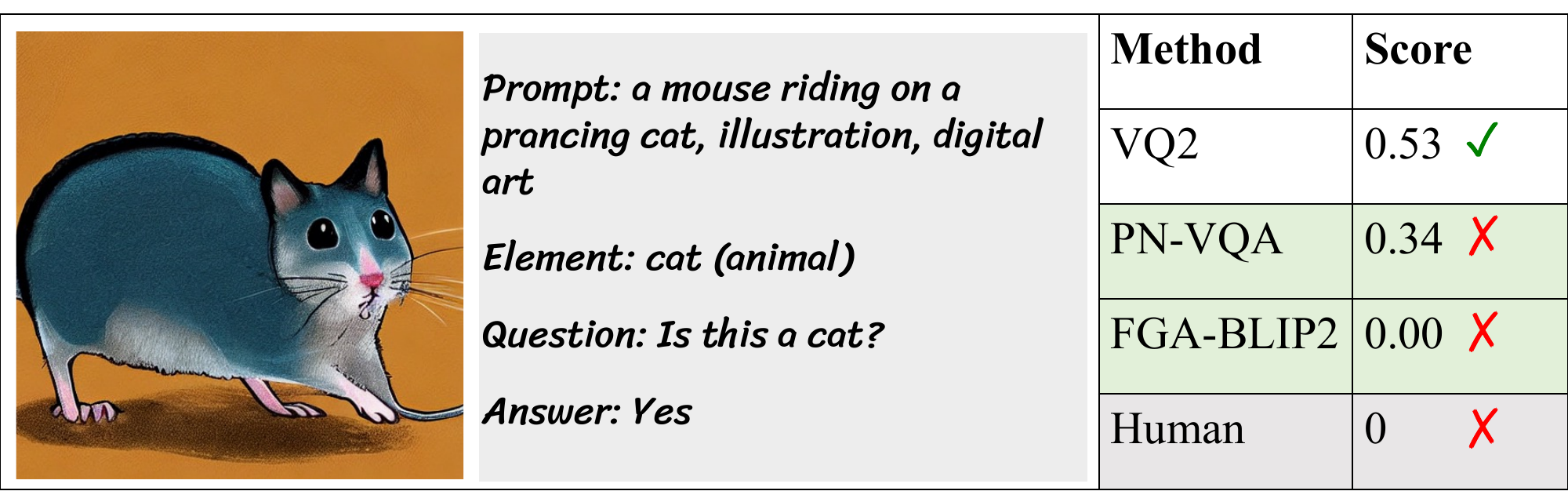}
    \end{minipage}
    \begin{minipage}{0.5\textwidth}
        \centering
        \includegraphics[width=\linewidth]{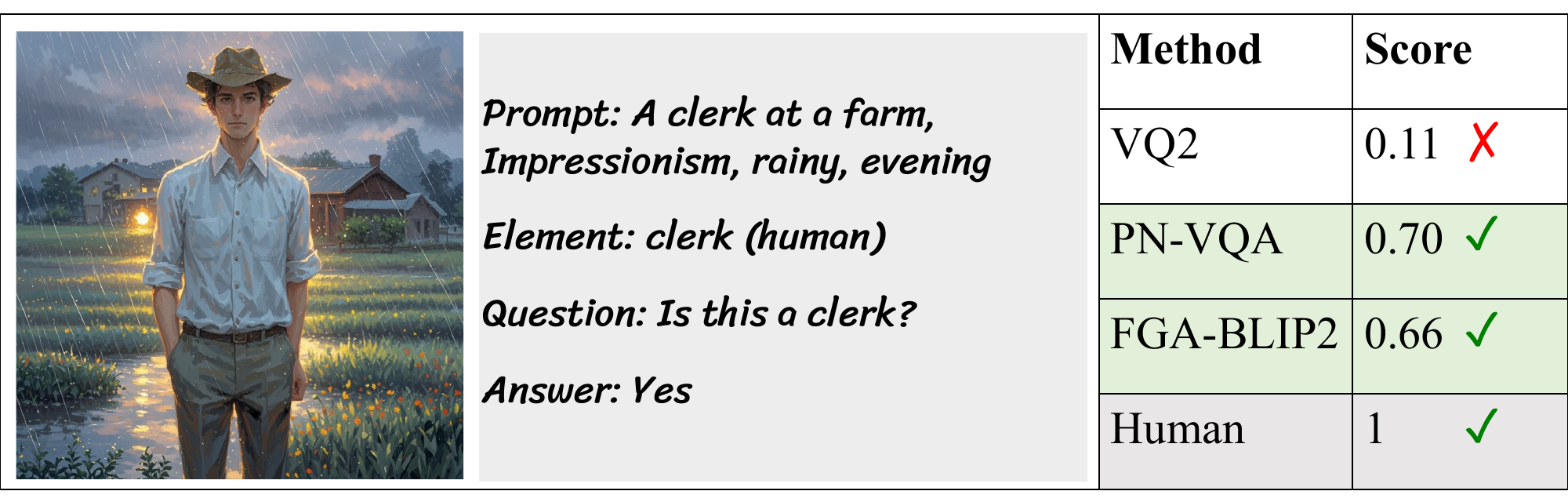}
    \end{minipage}
    \vspace{-2mm}
    \caption{Visualization of fine-grained alignment scores from different methods. In the figure, the question and answer are generated based on the element in the prompt and the score represents the probability of the element in the prompt matching the image (\cmark indicates a match, and \xmark indicates a mismatch). Here, both VQ2~\cite{yarom2024you} and PN-VQA use Qwen2-VL~\cite{wang2024qwen2} for VQA task. ``Human" refers to the average fine-grained scores from multiple annotators. It can be seen that our FGA-BLIP2 and PN-VQA yield results consistent with human annotations.}
    \label{fig:vis_ele_supp}
    \vspace{-2mm}
\end{figure*}

\subsection{Results of Metrics on Specific Skills}\label{sec:skill_eval}
We have categorized elements by different skills when annotating elements at a fine-grained level. Here, we analyze the accuracy of our fine-grained evaluation metrics on specific skills. The specific results are shown in Tab.~\ref{tab:acc_fine}. Vertically, our method achieves relatively high accuracy across different categories.  Horizontally, it can be observed that there is a strong correlation between the skills that the AIGC model performs poorly and the skills that the evaluation model does not assess accurately.
Notably, the counting category is an exception: while text-to-image models often make quantity errors in generated images, evaluating whether the image and text align in terms of quantity is comparatively easier.

\subsection{More Qualitative Comparisons}\label{sec:more_vis}
As shown in Figs.~\ref{fig:vis_score_supp} and~\ref{fig:vis_ele_supp}, we present additional visual results to illustrate that our method provides improved alignment evaluation between generated images and prompts.

\subsection{Structural Problem Evaluation}\label{sec:struct}

\begin{table*}[t]
    \centering
    \caption{Prompts for Structural Problem Evaluation. For different types of structural problems, we design specific question prompts and provided them to MLLMs to evaluate whether the generated images exhibit structural problems.}
    
    \resizebox{\linewidth}{!}{
        \begin{tabular}{ccl}
        \toprule
        \textbf{Category}  & \textbf{Fine-grained} & \textbf{Prompt}\\
        \midrule
        \multirow{2}[2]{*}{Animal} & \multirow{2}[2]{*}{-} & do the limbs, bodies, and faces of the animals in the image not conform to\\&& the laws of reality in terms of structure\\
        \midrule
        \multirow{2}[2]{*}{Object} & \multirow{2}[2]{*}{-} & do the objects in the image not conform to the laws of reality physically,\\&& including but not limited to the position and shape of the objects \\
        \midrule
        \multirow{2}[2]{*}{Human\ Body} &
        \multirow{2}[2]{*}{-} & do the limbs, bodies, and faces of the people in the image not conform to\\&& the laws of reality in terms of structure\\
        \midrule
        \multirow{9}[2]{*}{Human\ Body} &
        Face-Missing/Extra\ Feature  & does the person in the image have any missing or redundant features on the face\\&
        Face-Distorted/Exaggerated&does the person in the image have distorted features or exaggerated proportions on the face\\&
        Limb-Missing/Extra\ Limbs & does the person in the image have extra or fewer limbs\\&
        Limb-Distorted/Deformed  &  does the person in the image have any limb distortion or deformity\\&
        Limb-Disproportionate  &  does the person in the image have disproportionate limbs\\&
        Palm-Shapeless  &  does the person in the image have a shapeless palm\\&
        Palm-Missing/Extra Finger & does the person in the image have extra or fewer fingers on the hand\\&
        Palm-Deformed  &  does the person in the image have deformed fingers on the hand\\&
        Palm-Overlapping & does the person in the image have multiple hands overlapping and confusing\\
        \bottomrule
        \end{tabular}
    }
    \label{tab:prompt_structural}
\end{table*}

EvalMuse-40K's annotation on structural problems allows us to measure the ability of MLLMs to identify structural problems in generated images.
Such ability can be reflected by the model's accuracy and recall.
The accuracy can reflect the correctness and confidence of the model's judgment, and recall can reflect the model's ability to check all questions.
We use the VQA method to pose judgment questions to the MLLM, then apply softmax to the logits of the model's outputs (``Yes.'' and ``No.'') to obtain the probability for each response.

\noindent \textbf{Prompts.}
We use a unified template as ``Does this generated image have any structural problems? Specifically, $<cls\_prompt>$? Please answer yes or no only.'' where $<cls\_prompt>$ varies according to different tests and different structural problem labels.
For the accuracy test, we only test on coarse-grained ``animal'', ``object'', and ``human-body'' labels.
For the recall test, we test all the annotated structural problems, including coarse-grained and fine-grained. 
The $<cls\_prompt>$ for testing are shown in Tab.~\ref{tab:prompt_structural}.

\noindent \textbf{Select Threshold.}
We define the threshold as the gate-value of the model answering ``Yes.'', which means if the probability is larger than the threshold, the model's output will be chosen ``Yes.''.
Since MLLMs have different tendencies to answer Yes or No, if we simply use 0.5 as the threshold for answering Yes or No, it will lead to unfair testing.
Therefore, before calculating the accuracy and recall, we first select an optimal threshold for each model.

We denote positive samples as image-text pairs with Ground Truth ``Yes.'' and negative samples as with Ground Truth ``No.''.
For each tested model, we calculate the accuracy of positive and negative samples for all potential threshold values (from 0 to 1, with an interval of 0.01) on all test data.
We calculate the F1-Score using the accuracy of positive and negative samples and select the value with the highest F1-Score as the best threshold for the model.

\noindent \textbf{Accuracy.} 
Considering that each model has different abilities to assess different categories of objects, we test all three coarse-grained categories for each image.
The probability of model output ``Yes.'' is compared to the threshold to obtain the final output of the model. 
For each category, the final accuracy rate is the average of accuracy of positive and negative samples.
The results of the accuracy test are shown in Tab.~\ref{tab:acc}. 
\begin{table}[htbp]
    \centering
    \caption{The accuracy of three models on three labels. The accuracy of these models on the object test is around 50\%, proving that the perception of these models deviates from human definition.}
    \resizebox{\linewidth}{!}{
    \begin{tabular}{l|c|ccc}
    \toprule
    Model                          &  Threshold  &       Animal       &      Object       &      Human         \\
    \midrule
    LLaVA1.6~\cite{liu2024llava}   &     0.56    &       81.78\%      &      51.07\%      &      67.03\%       \\
    mPLUG-Owl3~\cite{ye2024mplug}  &     0.25    &  \textbf{86.46\%}  &      52.41\%      & \textbf{79.50\%}   \\
    Qwen2-VL~\cite{wang2024qwen2}  &     0.47    &       82.34\%      &  \textbf{52.83\%} &      75.28\%       \\
    \bottomrule
    \end{tabular}
    }

    \label{tab:acc}
\end{table}

\noindent \textbf{Recall.} 
We use all the images with structural problems annotated and the prompts we have shown in Tab.~\ref{tab:prompt_structural} to ask MLLMs questions.
Similar to the Accuracy test, the probability of model output ``Yes.'' is compared to the threshold to obtain the final output of the model. 
We count the number of times that MLLM answers ``Yes.'' and divide it by the total number of questions to get the recall rate of the model.

For example, if a picture is labeled with ``Object", ``Human\ Body-Face-Missing/Extra\ Feature".
We will ask the MLLM twice. 
The first time, we ask the model whether the picture has the structural problem of object, and the second time, we ask whether the picture has the structural problem of missing or extra features on the face.
The results of the recall test are shown in Tab.~\ref{tab:recall}.
\begin{table}[!tbp]
    \centering
    \caption{The recall rates of the three models. The recall rates of the three models are relatively low, indicating that the ability of MLLMs to detect structural problems in images still requires significant improvement.}
    \resizebox{\linewidth}{!}{
    \begin{tabular}{l|c|c|c}
    \toprule
    Model         &  LLaVA1.6~\cite{liu2024llava}  &  mPLUG-Owl3~\cite{ye2024mplug}  &  Qwen2-VL~\cite{wang2024qwen2}   \\
    \midrule
    Threshold     &             0.56               &              0.25               &             0.47                \\
    Recall        &            41.53\%             &         \textbf{61.57\%}        &            35.09\%              \\
    \bottomrule
    \end{tabular}
    }
    \label{tab:recall}
\end{table}

\setlength{\tabcolsep}{2pt} 
\begin{table*}[htbp]
  \centering
  \caption{Evaluation of image-text alignment across different T2I models. The table reports the overall image-text alignment scores and fine-grained alignment scores for various skills, evaluated using FGA-BLIP2. Here, a./h. is abbreviation for animal/human.}
  \resizebox{\linewidth}{!}{
    \begin{tabular}{lcccccccccccc}
    \toprule
    model & overall\ score & attribute & location & color & object & material & a./h. & food  & shape & activity & spatial & counting \\
    \midrule
    Dreamina v2.0Pro~\cite{dreamina}        & $3.74_{(1)}$ & $0.821_{(1)}$ & $0.793_{(1)}$ & $0.706_{(1)}$ & $0.747_{(1)}$ & $0.689_{(2)}$ & $0.756_{(1)}$ & $0.700_{(4)}$ & $0.580_{(19)}$ & $0.662_{(1)}$ & $0.747_{(2)}$ & $0.477_{(1)}$ \\
    DALLE 3~\cite{ramesh2022hierarchical}   & $3.63_{(2)}$ & $0.814_{(3)}$ & $0.782_{(3)}$ & $0.692_{(2)}$ & $0.732_{(2)}$ & $0.701_{(1)}$ & $0.734_{(2)}$ & $0.700_{(4)}$ & $0.682_{(6)}$ & $0.644_{(2)}$ & $0.768_{(1)}$ & $0.438_{(2)}$ \\
    FLUX 1.1~\cite{flux}                    & $3.47_{(3)}$ & $0.819_{(2)}$ & $0.758_{(4)}$ & $0.660_{(3)}$ & $0.694_{(4)}$ & $0.638_{(3)}$ & $0.686_{(3)}$ & $0.673_{(11)}$ & $0.607_{(17)}$ & $0.596_{(4)}$ & $0.671_{(4)}$ & $0.362_{(3)}$ \\
    Midjourney v6.1~\cite{midjourney}       & $3.33_{(4)}$ & $0.807_{(5)}$ & $0.736_{(8)}$ & $0.637_{(4)}$ & $0.693_{(5)}$ & $0.625_{(4)}$ & $0.619_{(4)}$ & $0.718_{(1)}$ & $0.659_{(13)}$ & $0.599_{(3)}$ & $0.716_{(3)}$ & $0.285_{(8)}$ \\
    SD 3~\cite{esser2024scaling}                                    & $3.27_{(5)}$ & $0.790_{(10)}$ & $0.728_{(11)}$ & $0.595_{(5)}$ & $0.695_{(3)}$ & $0.546_{(5)}$ & $0.560_{(10)}$ & $0.716_{(2)}$ & $0.637_{(14)}$ & $0.559_{(8)}$ & $0.646_{(7)}$ & $0.305_{(5)}$ \\
    Playground v2.5~\cite{li2024playground} & $3.20_{(6)}$ & $0.812_{(4)}$ & $0.785_{(2)}$ & $0.544_{(8)}$ & $0.657_{(8)}$ & $0.541_{(6)}$ & $0.578_{(6)}$ & $0.709_{(3)}$ & $0.675_{(7)}$ & $0.574_{(6)}$ & $0.634_{(10)}$ & $0.262_{(13)}$ \\
    SDXL-Turbo~\cite{sauer2023adversarial}  & $3.15_{(7)}$ & $0.788_{(12)}$ & $0.714_{(16)}$ & $0.494_{(12)}$ & $0.659_{(7)}$ & $0.487_{(10)}$ & $0.567_{(8)}$ & $0.671_{(12)}$ & $0.665_{(10)}$ & $0.551_{(9)}$ & $0.644_{(8)}$ & $0.306_{(4)}$ \\
    HunyuanDiT~\cite{li2024hunyuan}         & $3.08_{(8)}$ & $0.794_{(7)}$ & $0.753_{(6)}$ & $0.555_{(7)}$ & $0.666_{(6)}$ & $0.524_{(8)}$ & $0.576_{(7)}$ & $0.682_{(8)}$ & $0.705_{(2)}$ & $0.586_{(5)}$ & $0.648_{(6)}$ & $0.247_{(14)}$ \\
    Kandinsky3~\cite{arkhipkin2023kandinsky}& $3.08_{(8)}$ & $0.793_{(8)}$ & $0.723_{(12)}$ & $0.541_{(9)}$ & $0.652_{(9)}$ & $0.513_{(9)}$ & $0.583_{(5)}$ & $0.681_{(9)}$ & $0.661_{(11)}$ & $0.564_{(7)}$ & $0.665_{(5)}$ & $0.291_{(7)}$ \\
    SDXL~\cite{lin2024sdxl}                 & $2.99_{(10)}$ & $0.786_{(14)}$ & $0.717_{(15)}$ & $0.467_{(15)}$ & $0.623_{(11)}$ & $0.463_{(14)}$ & $0.533_{(12)}$ & $0.677_{(10)}$ & $0.660_{(12)}$ & $0.531_{(11)}$ & $0.607_{(12)}$ & $0.276_{(9)}$ \\
    PixArt-$\Sigma$~\cite{chen2024pixart}   & $2.98_{(11)}$ & $0.792_{(9)}$ & $0.755_{(5)}$ & $0.564_{(6)}$ & $0.633_{(10)}$ & $0.533_{(7)}$ & $0.561_{(9)}$ & $0.692_{(6)}$ & $0.703_{(3)}$ & $0.533_{(10)}$ & $0.641_{(9)}$ & $0.238_{(17)}$ \\
    Kolors~\cite{kolors}                    & $2.93_{(12)}$ & $0.790_{(10)}$ & $0.722_{(13)}$ & $0.498_{(11)}$ & $0.622_{(12)}$ & $0.480_{(11)}$ & $0.527_{(13)}$ & $0.621_{(15)}$ & $0.713_{(1)}$ & $0.496_{(16)}$ & $0.594_{(14)}$ & $0.245_{(15)}$ \\
    SDXL-Lightning~\cite{lin2024sdxl}       & $2.93_{(12)}$ & $0.788_{(12)}$ & $0.729_{(10)}$ & $0.478_{(14)}$ & $0.619_{(13)}$ & $0.458_{(15)}$ & $0.534_{(11)}$ & $0.619_{(16)}$ & $0.600_{(18)}$ & $0.528_{(12)}$ & $0.609_{(11)}$ & $0.274_{(10)}$ \\
    SSD1B~\cite{luo2023latent}              & $2.93_{(12)}$ & $0.798_{(6)}$ & $0.730_{(9)}$ & $0.502_{(10)}$ & $0.610_{(14)}$ & $0.480_{(11)}$ & $0.504_{(16)}$ & $0.688_{(7)}$ & $0.684_{(5)}$ & $0.508_{(14)}$ & $0.590_{(15)}$ & $0.297_{(6)}$ \\
    PixArt-$\alpha$~\cite{chen2023pixart}   & $2.88_{(15)}$ & $0.780_{(15)}$ & $0.738_{(7)}$ & $0.483_{(13)}$ & $0.607_{(15)}$ & $0.472_{(13)}$ & $0.521_{(14)}$ & $0.627_{(14)}$ & $0.670_{(9)}$ & $0.523_{(13)}$ & $0.600_{(13)}$ & $0.240_{(16)}$ \\
    IF~\cite{deep2023if}                    & $2.77_{(16)}$ & $0.725_{(19)}$ & $0.620_{(19)}$ & $0.452_{(18)}$ & $0.577_{(16)}$ & $0.416_{(18)}$ & $0.475_{(18)}$ & $0.570_{(18)}$ & $0.632_{(15)}$ & $0.498_{(15)}$ & $0.581_{(17)}$ & $0.188_{(20)}$ \\
    LCM-SDXL~\cite{luo2023latent}           & $2.77_{(16)}$ & $0.762_{(17)}$ & $0.706_{(17)}$ & $0.465_{(16)}$ & $0.575_{(17)}$ & $0.454_{(16)}$ & $0.513_{(15)}$ & $0.616_{(17)}$ & $0.615_{(16)}$ & $0.496_{(16)}$ & $0.587_{(16)}$ & $0.273_{(11)}$ \\
    PixArt-$\delta$~\cite{chen2024pixartdelta}& $2.73_{(18)}$ & $0.768_{(16)}$ & $0.718_{(14)}$ & $0.455_{(17)}$ & $0.565_{(18)}$ & $0.432_{(17)}$ & $0.486_{(17)}$ & $0.634_{(13)}$ & $0.685_{(4)}$ & $0.496_{(16)}$ & $0.574_{(18)}$ & $0.207_{(18)}$ \\
    LCM-SSD1B~\cite{luo2023latent}          & $2.66_{(19)}$ & $0.761_{(18)}$ & $0.683_{(18)}$ & $0.451_{(19)}$ & $0.540_{(19)}$ & $0.393_{(19)}$ & $0.457_{(19)}$ & $0.523_{(20)}$ & $0.673_{(8)}$ & $0.459_{(19)}$ & $0.572_{(19)}$ & $0.265_{(12)}$ \\
    SD v2.1~\cite{rombach2022high}          & $2.42_{(20)}$ & $0.698_{(20)}$ & $0.590_{(20)}$ & $0.354_{(20)}$ & $0.502_{(20)}$ & $0.363_{(21)}$ & $0.431_{(20)}$ & $0.532_{(19)}$ & $0.559_{(20)}$ & $0.398_{(20)}$ & $0.528_{(20)}$ & $0.190_{(19)}$ \\
    SD v1.5~\cite{rombach2022high}          & $2.25_{(21)}$ & $0.671_{(21)}$ & $0.534_{(21)}$ & $0.328_{(21)}$ & $0.470_{(22)}$ & $0.337_{(22)}$ & $0.372_{(22)}$ & $0.487_{(22)}$ & $0.500_{(22)}$ & $0.352_{(21)}$ & $0.488_{(22)}$ & $0.180_{(22)}$ \\
    SD v1.2~\cite{rombach2022high}          & $2.25_{(21)}$ & $0.659_{(22)}$ & $0.515_{(22)}$ & $0.315_{(22)}$ & $0.471_{(21)}$ & $0.377_{(20)}$ & $0.393_{(21)}$ & $0.498_{(21)}$ & $0.547_{(21)}$ & $0.349_{(22)}$ & $0.493_{(21)}$ & $0.181_{(21)}$ \\
    \bottomrule
    \end{tabular}%
    }
  \label{tab:t2i}%
\end{table*}%

\section{T2I Model Alignment Evaluation}\label{sec:t2i_eval}
Benefiting from the strong correlation between FGA-BLIP2 and human preferences in image-text alignment evaluation, we employ it to assess and rank the performance of widely used T2I models.

\noindent \textbf{Data Collection.} Considering that the variance of prompts calculated during FGA-BLIP2 training can reflect the potential of a prompt to generate diverse images with varying alignment scores, we prioritize selecting prompts with higher variance. This approach better distinguishes the image-text alignment capabilities of different models. 
We select the top 500 prompts with the highest variance from both the real prompts and synthesized prompts. 
To ensure diversity in prompt categories, we then apply the sampling strategy mentioned in Section~\ref{milp} to downsample the prompts to 100 from each group, resulting in a final set of 200 prompts. 
Subsequently, each T2I model generated four distinct images for each prompt to ensure the robustness of the evaluation.

\noindent\textbf{Evaluation Results.}
We use FGA-BLIP2 to evaluate both the overall alignment scores and the fine-grained scores between the images generated by different models and the corresponding text prompts. 
The detailed evaluation results are presented in Tab.~\ref{tab:t2i}.
It can be observed that proprietary models, such as Dreamina v2.0pro, DALLE 3, and FLUX 1.1, generally perform better in image-text alignment.

\end{document}